\documentclass[journal]{IEEEtran}
\ifCLASSINFOpdf
\else
\fi
\usepackage{graphicx}
\usepackage{amsmath}
\usepackage{stfloats}
\usepackage{multirow}
\usepackage{color}
\usepackage{booktabs}
\usepackage{CJKutf8}
\usepackage{pifont}

\usepackage{amsthm, bm,xspace}
\usepackage{amsfonts}
\usepackage{algorithm}
\usepackage{algorithmic}
\usepackage[algo2e]{algorithm2e}
\usepackage{subcaption}

\newcommand{\eg}{\emph{e.g.}}
\newcommand{\ie}{\emph{i.e.}}

\newcommand{\bx}{\mathbf{x}}
\newcommand{\pd}{\mathrm{d}}

\begin{document}
% \title{Learning from Decomposition-Fusion Feature for Blind Imgage Super-resolution}

\title{cDPMSR: Conditional Diffusion Models for Single Image Super-Resolution}
\title{A simple framework for Single Image Super-resolution via Accelerated Conditional Diffusion  Models}
\title{ACDMSR: Accelerated Conditional Diffusion Models for Single Image Super-Resolution}

\author{
Axi Niu, Pham Xuan Trung, Kang Zhang, Jinqiu Sun*, Yu Zhu, In So Kweon,~\IEEEmembership{Member, ~IEEE}, \\and Yanning Zhang,~\IEEEmembership{Senior Member,~IEEE}
%         %John~Doe,~\IEEEmembership{Fellow,~OSA,}
%         %and~Jane~Doe,~\IEEEmembership{Life~Fellow,~IEEE}% <-this % stops a space

% \author{ ~\IEEEmembership{Member, ~IEEE}, ~\IEEEmembership{Senior Member,~IEEE}
        %John~Doe,~\IEEEmembership{Fellow,~OSA,}
        %and~Jane~Doe,~\IEEEmembership{Life~Fellow,~IEEE}% <-this % stops a space
\thanks{This work is funded in part by the Project of the National Natural Science Foundation of China under Grant 61871328, Natural Science Basic Research Program of Shaanxi under Grant 2021JCW-03, as well as the Joint Funds of the National Natural Science Foundation of China under Grant U19B2037.). (* Corresponding author:Jinqiu Sun.)

Axi Niu, Yu Zhu, and Yanning Zhang are with the School of Computer Science, Northwestern Polytechnical University, Xi’an, 710072, China, and also with the National Engineering Laboratory for Integrated Aero-Space-Ground-Ocean Big Data Application Technology, Xi’an, 710072, China (email: nax@mail.nwpu.edu.cn, yuzhu@mail.nwpu.edu.cn, ynzhang@nwpu.edu.cn )

Pham Xuan Trung, Kangzhang, and In So Kweon are with the School of Electrical Engineering, Korea Advanced Institute of Science and Technology. (email:trungpx@kaist.ac.kr, kangzhang@kaist.ac.kr, iskweon77@kaist.ac.kr)

Jinqiu Sun is with the School of Astronautics, Northwestern Polytechnical University, Xi’an 710072, China (email: sunjinqiu@nwpu.edu.cn)
}
}

%M. Shell was with the Department of Electrical and Computer Engineering, Georgia Institute of Technology, Atlanta,GA, 30332 USA e-mail: (see http://www.michaelshell.org/contact.html).}% <-this % stops a space
% \thanks{J. Doe and J. Doe are with Anonymous University.}% <-this % stops a space
% \thanks{Manuscript received April 19, 2005; revised August 26, 2015.}}


\markboth{Journal of \LaTeX\ Class Files,~Vol.~14, No.~8, August~2015}%
{Shell \MakeLowercase{\textit{et al.}}: Bare Demo of IEEEtran.cls for IEEE Journals}



\maketitle

\begin{abstract}
Diffusion models have gained significant popularity in the field of image-to-image translation. Previous efforts applying diffusion models to image super-resolution (SR) have demonstrated that iteratively refining pure Gaussian noise using a U-Net architecture trained on denoising at various noise levels can yield satisfactory high-resolution images from low-resolution inputs. However, this iterative refinement process comes with the drawback of low inference speed, which strongly limits its applications. To speed up inference and further enhance the performance, our research revisits diffusion models in image super-resolution and proposes a straightforward yet significant diffusion model-based super-resolution method called ACDMSR (accelerated conditional diffusion model for image super-resolution). Specifically, our method adapts the standard diffusion model to perform super-resolution through a deterministic iterative denoising process. Our study also highlights the effectiveness of using a pre-trained SR model to provide the conditional image of the given low-resolution (LR) image to achieve superior high-resolution results. 
We demonstrate that our method surpasses previous attempts in qualitative and quantitative results through extensive experiments conducted on benchmark datasets such as Set5, Set14, Urban100, BSD100, and Manga109. Moreover, our approach generates more visually realistic counterparts for low-resolution images, emphasizing its effectiveness in practical scenarios.
\end{abstract}

\begin{IEEEkeywords}
Diffusion Models, Image-to-Image Translation,  Conditional Image Generation, Image Super-resolution.
\end{IEEEkeywords}

\IEEEpeerreviewmaketitle

\section{Introduction}
\label{sec:intro}
\IEEEPARstart{S}{ingle image super-resolution (SISR)} has drawn active attention due to its wide applications in computer vision, such as object recognition, remote sensing and so on~\cite{9381876,chen2022consistent,pham2022self,niu2022fast,pan2020unsupervised,pan2022ml,pan2022labeling,chang2021two}. 
SISR aims to obtain a high-resolution (HR) image containing great details and textures from a low-resolution (LR) image by a super-resolution method, which is a classic ill-posed inverse problem~\cite{niu2022ms2net,niu2023cdpmsr}. 
To establish the mapping between HR and LR images, lots of CNN-based methods have emerged~\cite{ledig2017photo,zhang2018image,ma2019matrix,cai2019toward,zuo2019multi,liu2020photo,lyn2020multi,li2021single,9897540}. 
These methods focus on designing novel architectures by adopting different network modules, such as residual blocks~\cite{lim2017enhanced,hu2019channel}, attention blocks~\cite{li2019filternet,niu2023gran}, non-local blocks~\cite{deng2015single,mei2021image}, transformer layers~\cite{liang2021swinir,lu2022transformer}, and contrastive learning~\cite{wu2021practical,zhu2022self,niu2023learning}. 
For optimizing the training process, they prefer to use the MAE or MSE loss (\eg, $L_{1}$ or $L_{2}$) 
to optimize the architectures, which often leads to over-smooth results because the above losses provide a straightforward learning objective and optimize for the popular PSNR (peak signal-to-noise-ratio) metric~\cite{blau2018perception,he2018cascaded,delbracio2020projected,freirich2021theory,whang2022deblurring}.

With deep generative models of all kinds exhibiting high-quality samples in a wide variety of data modalities, approaches based on the deep generative model have become one of the mainstream, mainly including GAN-based methods~\cite{wang2018esrgan,soh2019natural,tian2022generative} and flow-based methods~\cite{lugmayr2020srflow,liang2021hierarchical,wolf2021deflow}, which have shown convincing image generation ability. 
GAN-based SISR methods~\cite{wang2018esrgan,soh2019natural,tian2022generative} often introduce a generator and a discriminator in an adversarial way to push the generator to generate realistic images. The generator can generate an SR result for the input LR, and the discriminator aims to distinguish if the generated SR result is true. The training process is optimized by combining content loss and adversarial losses, which have strong learning abilities~\cite{tian2022generative,bell2019blind,emad2021dualsr}. While GAN-based methods have an obvious drawback in that they easily fall into mode collapse, the training process is challenging to converge with complex optimization~\cite{metz2016unrolled,ravuri2019classification,huang2020fast}. Furthermore,  adversarial losses often introduce artifacts not present in the original clean image, leading
to large distortion~\cite{lugmayr2021ntire,whang2022deblurring}. Flow-based SR methods are another famous line based on the deep generative model. They directly account for the ill-posed problem with an invertible encoder~\cite{saharia2022image,laroche2022bridging}. The flow-based operation transforms a Gaussian distribution into an HR image space instead of modeling one single output and inherently resolves the pathology of the original ”one-to-many” SR problem. Optimized by a negative loglikelihood loss, these methods avoid training instability. Still, they suffer from enormous footprints and high training costs due to the strong architectural constraints to keep the bijection between latents and data~\cite{saharia2022image}.

Lately, the broad adoption of diffusion models has shown promising results in image generative tasks~\cite{ho2020denoising}.
In SRDiff~\cite{li2022srdiff}, the authors propose a two-stage SR framework. First, they design a super-resolution structure and pre-train it to obtain a conditional image for the diffusion process. Then they redesign the U-net structure in diffusion models. The training process of this method is relatively complicated, and it does not consider combining existing pre-trained SR models, such as EDSR~\cite{lim2017enhanced}, RCAN~\cite{zhang2018image}, and SwinIR~\cite{liang2021swinir}. Similarly, SR3~\cite{saharia2022image} directly applies the bicubic up-sampled LR image as the conditional image. Nevertheless, the stochastic sampling style in the inference phase makes the reconstruction process complex and slow.
Unlike them, we propose a simple but non-trivial method for image super-resolution based on the conditional diffusion model,\ie, ACDMSR (accelerated conditional diffusion model for image super-resolution). Our work shares some similarities with SRDiff, which first applies diffusion models to the SR tasks. Different from the existing technique~\cite{li2022srdiff,saharia2022image}, our ACDMSR adopts the current pertained SR methods to provide the conditional image, which is more plausible than the one in~\cite{saharia2022image,li2022srdiff}. 
Moreover, it helps to significantly improve perceptual quality over existing state-of-the-art methods across multiple standard benchmarks. 
Furthermore, to accelerate the inference steps, we build a $n$-th order sampler that decreases the 1000-step- to 40-step inference and keeps the good quality. Compared with previous diffusion model-based methods SR3 and SRDiff, which need 1000 inference steps, ours significantly shortened the acquisition of final results.
By simply concatenating a Gaussian noise and the conditional image with $L_{1}$ loss optimizing the diffusion model, our method makes the training process more concise compared with ~\cite{saharia2022image,li2022srdiff}. The main contributions of this work are listed as follows:

\begin{itemize}

    \item To the best of our knowledge, we are the first to combine diffusion models and the existing pre-trained SR models to conduct image super-resolution, which can also be taken as a post-process framework.

    \item Compared with existing diffusion model-based SR methods, our ACDMSR adopts a deterministic sampling way in the inference phase. It can effectively reduce the inference steps from 1000 to just 40, achieving an improved equilibrium between distortion and perceptual quality.

    \item Compared to existing SOTA SR methods, our ACDMSR achieves superior perceptive results and can generate more photo-realistic SR results on various benchmarks.
    
\end{itemize}

\section{Related Work}
\subsection{Single Image Super-resolution Methods}
\textbf{CNN-based methods.} 
CNN-based methods are a trendy line for image super-resolution, and much great work is coming out. For example,~\cite{ledig2017photo} employs the ResNet architecture from~\cite{he2016deep} and solves the time and memory issues with good performance. Then~\cite{lim2017enhanced} further optimizes it by analyzing and removing unnecessary modules to simplify the network architecture and produce better results. After them, RCAN~\cite{zhang2018image} and MCAN~\cite{ma2019matrix}, and EMASRN~\cite{zhu2021lightweight} adopt the attention mechanism~\cite{woo2018cbam} and design new residual dense networks. Then MLRN ~\cite{lyn2020multi}, SRNIF~\cite{li2021single}, and BSRT~\cite{li2022blueprint} proposed multi-scale fusion or internal and external features fusion architecture to solve the problem that the existing SISR could not make full use of the characteristic information of the middle network layer and internal features. In addition, SwinIR~\cite{liang2021swinir} and ESRT~\cite{lu2022transformer} apply transformer technology to improve the performance further. While these methods aim at pursuing higher PSNR (peak signal-to-noise-ratio) by designing novel architectures and using the MSE or MAE loss (\eg, $L_{1}$ or $L_{2}$) to optimize the architectures, which often leads to smooth results because the above losses provide a straightforward learning objective~\cite{blau2018perception,delbracio2020projected,freirich2021theory,whang2022deblurring}.

\textbf{Generative model-based methods.}

Because deep generative models have recently exhibited promising results in generating images with rich details, it has become popular to adopt generative models to conduct image super-resolution, such as  GAN-based methods~\cite{ledig2017photo,wang2018esrgan,soh2019natural,tian2022generative} and flow-based methods~\cite{lugmayr2020srflow,liang2021hierarchical,wolf2021deflow}.
SRGAN~\cite{ledig2017photo} is the first GAN-based SISR method. It adopts the GAN technology to push the generator to produce results with better Visual effects. Compared with SRGAN,   ESRAGN~\cite{wang2018esrgan} trains the discriminator to predict the authenticity of the generated image instead of predicting if the generated image is valid. NatSR~\cite{soh2019natural} proposes a Naturalness Loss based on a pre-trained natural manifold discriminator to improve the ability of the discriminator and achieve comparable results to recent CNNs. 
However, GAN-based methods have an obvious drawback that is jointly optimizing the whole training process by combining MAE or MSE  makes the model easy to fall into mode collapse, and the training process is not easy to converge with complex optimization~\cite{metz2016unrolled,ravuri2019classification}.
Furthermore,  adversarial losses often introduce artifacts not present in the original clean image, leading to large distortion~\cite{lugmayr2021ntire,whang2022deblurring}. 
Flow-based SR methods are another famous line based on the deep generative model. They directly account for the ill-posed problem with an invertible encoder~\cite{saharia2022image,laroche2022bridging}. The flow-based operation transforms a Gaussian distribution into an HR image space instead of modeling one single output and inherently resolves the pathology of the original ”one-to-many” SR problem. Optimized by a negative loglikelihood loss, these methods avoid training instability. Still, they suffer from enormous footprints and high training costs due to the strong architectural constraints to keep the bijection between latents and data~\cite{saharia2022image}.

\subsection{Diffusion Models}
Diffusion models have achieved promising results in image generation~\cite{ho2020denoising,song2020denoising,bansal2022cold}. It aims to use a Markov chain to transform latent variables in simple distributions (e.g., Gaussian) to data in complex distributions. The core technology for the success of diffusion models is their iterative sampling process. It progressively removes noise from a random noise vector. This iterative refinement procedure repetitively evaluates the diffusion model, allowing for the trade-off of compute for sample quality: by using extra compute for more iterations, a small-sized model can unroll into a larger computational graph and generate higher quality samples~\cite{karras2022elucidating,salimans2022progressive,song2023consistency}.
Inspired by the above works, some researchers apply diffusion models in low-level vision tasks~\cite{whang2022deblurring,li2022srdiff,saharia2022image}. In~\cite{whang2022deblurring}, authors propose a novel framework for blind image deblurring based on conditional diffusion models, which employs a stochastic sampler to refine the output of a deterministic predictor and produces a diverse set of plausible reconstructions for a given input, leading to a significant improvement in perceptual quality over existing state-of-the-art methods. SRdiff~\cite{li2022srdiff} also discuss the drawbacks of current generative models-based SR methods. It designs a novel single-image super-resolution model based on diffusion models, which can provide diverse and realistic super-resolution predictions while avoiding issues with over-smoothing, mode collapse, or large model footprints. At the same time, it has to combine the counterpart output from the pre-trained SR model for the LR input, which makes the whole training process and forward diffusion process very complex and may struggle with images that contain complex textures or patterns. Unlike SRdiff, SR3~\cite{saharia2022image} presents a straightforward style to introduce diffusion models to help image super-resolution. It just takes the bicubic low-resolution image as the conditional image and uses denoising diffusion probabilistic models to perform stochastic denoising and achieve super-resolution through iterative refinement using a U-Net model trained on denoising at various noise levels, achieving strong performance on super-resolution tasks on faces and natural images, as well as effective cascaded image generation. 
Though these methods have achieved plausible visual quality, they have an obvious drawback: the sampling speed needs to be improved in the inference time.

\section{Perliminaries: Overview of Diffusion Models}
In diffusion models, a Markov chain of diffusion steps generates data by progressively perturbing the data with Gaussian noise.
Subsequently, these models aim to learn how to reverse the diffusion process and reconstruct desired data samples from the noise. This section begins by revisiting the standard denoising diffusion probabilistic model (DDPM)~\cite{ho2020denoising} to provide a basic understanding. A typical probabilistic diffusion model consists of four main components: the forward process, the reverse process, the optimization of the diffusion model, and the inference stage. We will now introduce each of these components in the following sections:

\subsection{Forward process}
Suppose we have a real data distribution $\bx_0\sim q(\bx)$.
The forward process gradually adds noise into a sampled image $\bx_0$ using a variance (noise) schedule $\beta_1,\ldots,\beta_T$ ($\beta_{t} \in (0,1),1\le t\le T$) to generate noised versions $\bx_1,\bx_2,\ldots,\bx_T$ from the original image $\bx_0$. This process can be defined with a Markovian structure:
\begin{equation}
\label{eq:ddpmforward_diffusion}
q(x_t|x_{t-1})=\mathcal{N}(x_t;\sqrt{1-\beta_t} \bx_{t-1}, \beta_t \mathbf{I}), \quad 1 \leq t \leq T.
\end{equation}
By leveraging the properties of the Gaussian distribution and marginalizing the intermediate steps, we can sample $\bx_t$ at any given time-step $t$ using the following formulation:
\begin{equation}
q(\bx_t|\bx_0)=\mathcal{N}(x_t; \sqrt{\hat\alpha_t} \bx_0,(1-\hat\alpha_t)\mathbf{I}),
\label{eq:ddpmtrans}
\end{equation}
where $\alpha_t=1-\beta_t$ and $\hat\alpha_t=\prod^t_{s=1}\alpha_t$. This formulation allows us to express $\bx_t$ using the reparameterization trick:
\begin{equation}
\bx_t(\bx_0, \epsilon)=\sqrt{\hat\alpha_t} \bx_0+\sqrt{1-\hat\alpha_t}\bm{\epsilon},
\label{equ:forwardx}
\end{equation}
where $\bm{\epsilon}$ is a Gaussian noise vector with $\bm{\epsilon} \sim \mathcal{N}(0, \mathbf{I})$.

\subsection{Reverse process}

In order to acquire a real sample $\bx_0$ from a Gaussian noise input $\bx_{T}\sim \mathcal N(0, \mathbf{I})$, the reversal of the preceding forward process is required. This involves the construction of the inverse of Eq.~\ref{eq:ddpmforward_diffusion} and the iterative reversal using $q(\bx_{t-1}|\bx_t)$. It is worth highlighting that if $\beta_{t}$ is sufficiently small, $q(\bx_{t-1}|\bx_t)$ will also follow a Gaussian distribution. However, estimating $q(\bx_{t-1}|\bx_t)$ presents a challenge as it requires the utilization of the complete dataset. Furthermore, when conditioned on $\bx_0$ it becomes tractable:

\begin{equation}
    p(\bx_{t-1}|\bx_t,\bx_0)=\mathcal{N}({\bx_{t-1};\mu_(\bx_t,x_0),\sigma(\bx_t,\bx_0)}),
\end{equation}
where $\mu(\bx_t,\bx_0):=\frac{\sqrt{\hat{\alpha}_{t-1}}\beta_t}{1-\hat{\alpha}_t}\bx_0+\frac{\sqrt{\alpha_t}(1-\hat{\alpha}_{t-1})}{1-\hat{\alpha}_t}\bx_t$ and $\sigma(\bx_t, \bx_0):=\frac{1-\hat{\alpha}_{t-1}}{1-\hat{\alpha}_t}\beta_t$. By substitution Eq. \ref{equ:forwardx}, $\bx_0={(\bx_t - \sqrt{1-\hat\alpha_t}\epsilon)/\sqrt{\hat\alpha_t}}$, into the $\mu(\bx_t,\bx_0)$, we can have 
\begin{equation}
    \mu(\bx_t,\bx_0)=\frac{1}{\sqrt{\alpha_t}}(\bx_t-\frac{1-\alpha_t}{\sqrt{1-\hat{\alpha}_t}}\epsilon).
\end{equation}
Following the choice of~\cite{ho2020denoising}, if we train a model $q_\theta(\bx_{t-1}|\bx_t)=\mathcal{N}(\mu_\theta(\bx_t,t),\sum_\theta(\bx_t,t)I)$ to learn the above reverse process, $p(\bx_{t-1}|\bx_t,\bx_0)$, and set $\sum_\theta(\bx_t,t)$ as $\sum_\theta(\bx_t,t))=\sigma(\bx_t, \bx_0)$, we can use network $f_\theta$ to predict the noise $\epsilon\approx f_{\theta}(\bx_t, t)$ so that the reverse process becomes learnable:
\begin{equation}
\begin{aligned}
    &q_\theta(\bx_{t-1}|\bx_t)=\\
    &\mathcal{N}(\frac{1}{\sqrt{\alpha_t}}(\bx_t-\frac{1-\alpha_t}{\sqrt{1-\hat{\alpha}_t}}f_\theta(\bx_t,t)),
\frac{1-\hat{\alpha}_{t-1}}{1-\hat{\alpha}_t}\beta_tI). \\
\end{aligned}
\label{eq:reverseprocess}
\end{equation}

\subsection{Optimize the diffusion model}
In \cite{ho2020denoising}, it has been demonstrated that reweighted evidence lower bound proves to be an effective loss function in practical applications:
\begin{equation}
    L(\theta) = \mathbb{E}_{t,\bx,\epsilon}\|f_\theta(\bx_t, t) - \epsilon \|^2,
\end{equation}
where the model learns to predict the added noise $\epsilon$. The pseudocode for the training is shown in the training part of Algorithm~\ref{alg:ddpm}.

\subsection{Inference}
After training, the inference becomes trivial now, since given the start point $\bx_T$, we can get the formulation of next step image $\bx_{t-1}$ with the reparametrization trick for Equation~\ref{eq:reverseprocess} as follows:
\begin{equation}
    \bx_{t-1} \longleftarrow \frac{1}{\sqrt{\alpha_t}}(\bx_t-\frac{1-\alpha_t}{\sqrt{1-\hat{\alpha}_t}}f_\theta(\bx_t, t)) + \sqrt{1-\hat{\alpha}_t}\epsilon_t,
\label{eq:ddpmsampling}
\end{equation}
where $\epsilon_t\sim\mathcal{N}(\textbf{0}, \textbf{I})$ is the random noise added in each denoise step. We can sample the final image $\bx_0$ by iteratively applying the above equation. The pseudocode for the inference is shown in the inference part of Algorithm~\ref{alg:ddpm}.

\begin{algorithm}[H]
  \caption{DDPM}
  \label{alg:ddpm}
  \small
\textbf{Input:} Dataset $D$, noise predictor $f_\theta$, noise schedule $\hat{\alpha}_t$, total timestep $T$ \\
{\textbf {Training:} train $f_\theta$}
  \begin{algorithmic}[1]
    \REPEAT
      \STATE $\bx_0 \sim D$
      \STATE $t \sim [1, ...,T]$
      \STATE ${\epsilon}\sim\mathcal{N}(\mathbf{0},\mathbf{I} )$
      \STATE Take a gradient descent step on
      \STATE $\qquad \nabla_\theta \|\epsilon - f_\theta(\sqrt{\hat{\alpha}_t}\bx_0+\sqrt{(1-\hat{\alpha_t})}\epsilon, t)  \|^2$ 
    \UNTIL{converged}
  \end{algorithmic}
\textbf{Inference:} sampling $\bx_0$
    \begin{algorithmic}[1]
    \STATE ${\bx_T}\sim\mathcal{N}(\mathbf{0},\mathbf{I} )$
    \FOR{$t=T,...,1$}
    \STATE ${\epsilon_t}\sim\mathcal{N}(\mathbf{0},\mathbf{I} )$, if $t>1$, else $\epsilon_t=\mathbf{0}$
    \STATE    $\bx_{t-1} = \frac{1}{\sqrt{\alpha_t}}(x_t-\frac{1-\alpha_t}{\sqrt{1-\hat{\alpha}_t}}f_\theta(\bx_t, t)) + {\beta_t}\epsilon_t$
    \ENDFOR.
    \end{algorithmic}
\end{algorithm}

\begin{figure*}[!ht]
\centering	
	\includegraphics[width=12cm]{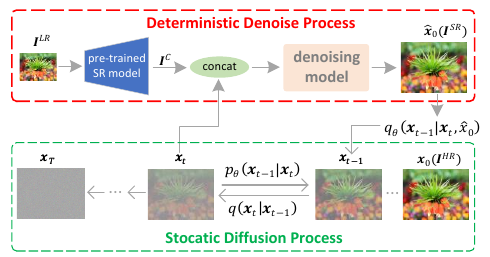}
	\caption{
	Illustration of our method. The model contains a stochastic forward diffusion process, gradually adding noise to an $\bm I^{HR}$ image. And a deterministic denoise process is applied to recover high-resolution and realistic images $\bm I^{SR}$  corresponding to $\bm I^{LR}$  images.}
	\label{fig:overview}
\end{figure*}

\section{Methodology}
Our method can be seen as a post-process for single image super-resolution (SISR).  
As shown in Fig.~\ref{fig:overview}, our ACDMSR consists of a stochastic diffusion process forward procedure that gradually adds noise to an image until a fully normal Gaussian noise and a deterministic denoising reverse process that conditions on $\bm I^C$ to reconstruct the image from noise. Algorithm~\ref{alg:cDMSR} shows the whole process of our ACDMSR. The following section introduces our method in detail.

\subsection{Stocatic Diffusion Process}
Given a SISR dataset $(\bm I^{HR}, \bm I^{LR})\sim D$, we adopt the diffusion model
~\cite{ho2020denoising,bansal2022cold} to map a normal Gaussian noise $\bx_T \sim\mathcal{N}(0, \bm 1)$ to a high-resolution image $\bx_0 = I^{HR}$ with a corresponding conditional image $\bx^C=I^{LR}$. We will talk about the choice of the conditional image later. The diffusion model contains latent variables $\bx=\{\bx_t|t=0,1,...,T\}$, where $\bx_0 = I^{HR}$, $ \bx_T = \mathcal{N}(0, \bm 1)$.
The same noise schedule with~\cite{ho2020denoising} is used for our method, $\beta_1, \ldots, \beta_T$ where $\quad 1 \leq t \leq T$.

\noindent\textbf{Forward stochastic diffusion process.} We define the forward process $q(\bx_t| \bm I^{HR}):=q(\bx_t| \bx_0)$ of diffusion model with a Gaussian process by the Markovian structure:

\begin{equation}
\label{eq:forward_diffusion}
\begin{aligned}
q(\bx_t|\bx_{t-1})=\mathcal{N}(\bx_t;\sqrt{1-\beta_t} \bx_{t-1}, \beta_t \mathbf{I}),\\
q(\bx_t|\bx_0)=\mathcal{N}(\bx_t; \sqrt{\hat\alpha_t} \bx_0,(1-\hat\alpha_t)\mathbf{I}).
\end{aligned}
\end{equation}

Same with DDPM~\cite{ho2020denoising}, the forward process gradually adds noise into an image $\bx_0$ to generate latent variables $\bx_1,...,\bx_T$ for the original image $\bx_0$. With the Gaussian distribution reparameterization trick, we can write the latent variable $\bx_t$ as Eq.~\ref{equ:forwardx}, $\bx_t(\bx_0, \epsilon)=\sqrt{\hat\alpha_t} \bx_0+\sqrt{1-\hat\alpha_t}\bm{\epsilon}$.

\noindent\textbf{Model training.} According to~\cite{nikankin2022sinfusion}, our findings demonstrate that predicting the image, rather than focusing on the noise, yields superior outcomes when applied in super-resolution tasks. We have proved it in Sec.~\ref{sec:ab}. Therefore, the optimization target of our diffusion model is denoising $\bx_t\sim p(\bx_t|\bx_0)$ to get estimated $ \hat{\bx}_0$ with a U-Net $f_\theta(\bx_t, t, \bx^C):=\hat{\bx}_0\approx \bx_0$. We use the following loss function to train the model:

\begin{equation}
     L := \mathbb E_{t,(\bx_0, I^C), \epsilon}[\| \bx_0 - f_\theta(\alpha_t \bx_0 + \sigma_t \bm\epsilon, t, \bx^C) \|^2],
\label{eq:cdpmsr_loss}
\end{equation}
where $t$ is uniformly sampled between 1 and $T$. With Eq.~\ref{equ:forwardx}, $\epsilon=(\bx_t-\sqrt{\hat\alpha_t} \hat{\bx}_0)/\sqrt{1-\hat\alpha_t}$, we can easily predict the added noise to the image $\bx_t$.

Here, different with~\cite{ho2020denoising}, we add an additional input $\bx^C$ as the conditional image to guide the model $f_\theta$ to keep the same content with $\bx^C$ during the denoising process. 

\begin{algorithm}[!htbp]
  \caption{ACDMSR}
  \label{alg:cDMSR}
\small
  \SetKwInOut{Training}{Training}\SetKwInOut{Inference}{Inference}
  \SetKwData{Input}{Input:}\SetKwFunction{Uniform}{Uniform}
  \SetKwData{Require}{Require:}
\Training {train denoising model $f_\theta$}
\Input Dataset $D$, schedule $\alpha_t,\sigma_t$, timesteps $T$, pre-trained super-resolution model $\phi_\theta$
  \BlankLine
  \begin{algorithmic}[1]
    \REPEAT
      \STATE $(\bm I^{HR}, \bm I^{LR}) \sim D$, $t \sim$ \Uniform{$\{1,...,T\}$}, \\${\bm \epsilon}\sim\mathcal{N}(\mathbf{0},\mathbf{I} )$
      \STATE $\bx_0=\bm I^{HR},\bx^{C} = \phi_\theta(I^{LR})$

      \STATE $\bx_t = \alpha_t\bm I^{HR}+\sigma_t\bm \epsilon$
      \STATE Take a gradient descent step on\\
       $\qquad \nabla_\theta \|\bm I^{HR} - f_\theta(\bx_t, t, \bx^{C})  \|^2$ 

    \UNTIL{converged} 
  \end{algorithmic}
\BlankLine

\Inference { super resolve $\bm I^{LR}$}
\Input trained denoising model $f_\theta$, pre-trained super-resolution model $\phi_\theta$, diffusion sampler $\mathcal{F}(f_\theta, \bx_t,t,\bx^C)$
\BlankLine
    \begin{algorithmic}[1]
    \STATE $\bx^{C} = \phi_\theta(I^{LR})$
    \STATE ${\bx_T}\sim\mathcal{N}(\bm{0},\bm{1} )$
    \FOR{$t=T,...,1$}
    \STATE $\bm \bx_{t-1} = \mathcal{F}(f_\theta, \bx_t,t,\bx^C)$
    \ENDFOR 
  \end{algorithmic}
\end{algorithm}

\subsection{Conditional image choice}
To get realistic super-resolution images, \cite{li2022srdiff, saharia2022image} also introduced diffusion models with conditional denoising on a pre-trained feature extractor or a bicubic upsampled image on a low-resolution image. In this work, we leverage the power of the current development of SISR to provide a better conditional image. Specifically, given a low-resolution image $I^{LR}$ and a pre-trained super-resolution model $\phi_\theta$, we generate our conditional image by $\bx^C=\phi_\theta(I^{LR})$, which has been proved to be more plausible for obtaining results with better perceptual quality in ablation study~\ref{ab:different_conditional_inputs}.

\subsection{Sampling Process}
Diffusion models are known to be slow and need thousands of forward evaluation steps to achieve the generated image with good quality. Similarly, the diffusion model-based super-resolution method inherited this drawback. To remedy this issue, we propose a $n$-th order sampler that adapts two accelerating sampling strategies from existing work, ~\ie, DDIM~\cite{song2020denoising} and DPM-solver~\cite{lu2022dpm}.
These two sampling strategy has been shown
to help the diffusion model achieve good image quality and keep a short sampling time.
In this section, we first define what a sampler is. Then we describe the proposed super-resolution sampler in detail.

\textbf{Iterative super-resolution sampler.} Given a pre-trained model $f_\theta$ with objective Eq~\ref{eq:cdpmsr_loss}, and a low-resolution conditional image $\bx^C$, we define a iterative super-resolution sampler from $t=T$ to $t=0$ as:
\begin{equation}
    \bx_{t-1} = \mathcal{F}(f_\theta,\bx_t,t,\bx^C)
\end{equation}
where $\bx_t$ is the ancestor of $\bx_{t-1}$.

\textbf{First order deterministic sampling.} Different from SR3 and SRdiff sampling via a stochastic way, we use a deterministic sampling method to conduct the iterative reverse process $\bx_{t-1} = \mathcal{F}(f_\theta,\bx_t,t,\bx^c)$ in a DDIM-like manner which has been shown achieve a high-quality image in limited inference steps. Given the image $\bx_t$ at step $t$, we can write the generation process of $\bx_{t-1}$ as follows:
\begin{equation}
\begin{aligned}
\bx_{t-1} &= \mathcal{F}_{\text{1st}}(f_\theta, \bx_t,t,\bx^c) \\
&= \sqrt{\hat\alpha_{t-1}}\hat{\bx}_0  + \sqrt{1-\hat{\alpha}_{t-1}}\frac{\bx_t - \sqrt{\hat\alpha_t}\hat{\bx}_0}{\sqrt{1-\hat\alpha_t}},
\label{eq:ddim}
\end{aligned}
\end{equation}
where $\hat{\bx}_0$ is predicted with trained denoising model $\hat{\bx}_0=f_\theta(\bx_t, t, \bx^C)$. Compared to the DDPM sampling process in Eq.~\ref{eq:ddpmsampling}, the above sampling does not add noise in each step, making it a deterministic method. Since, in each step, we need only one forward model evaluation, we call this method a first-order method.

\textbf{Second order deterministic sampling.} \cite{kingma2021variational} view the diffusion model as a stochastic differential equation (SDE), which has the same transition distribution $q(\bx_t|\bx_0)$ as in Eq~\ref{eq:ddpmtrans} for any $t\in[0,T]$:
\begin{equation}
    \pd\bx_t = f(t)\bx_t \pd t + g(t)\pd\bm w_t,\quad \bx_0\sim q_0(\bx_0),
    \label{eq:dpmsde}
\end{equation}
where $\bm w_t\in\mathbb{R}^D$ is the standard Wiener process, and
\begin{equation}
    f(t) = \frac{\pd\log{\sqrt{\hat\alpha_t}}}{\pd t},g(t)=\frac{1-\hat\alpha_t}{\pd t}-2f(t)\sqrt{\hat\alpha_t}.
\end{equation}

With some regularity, \cite{song2020score} shows that the above forward SDE Eq.\ref{eq:dpmsde} has an quivalent reverse process starting from the marginal distribution $q({\bx_T})$ at time $T$ to time step 0: 
\begin{equation}
    \frac{\pd\bx_t}{\pd t}=f(t)\bx_t-\frac{1}{2}g^2(t)\nabla_\bx\log q(\bx_t),\quad \bx_T\sim \mathcal{N}(0,\bm I),
\end{equation}
where score function $\nabla_\bx\log q(\bx_t)$ can be replaced with the noise prediction of a model $\epsilon_\theta(\bx_t,t)$, such that:
\begin{equation}
    \frac{\pd\bx_t}{\pd t}=f(t)\bx_t-\frac{g^2(t)}{2\sqrt{1-\hat\alpha_t}}\epsilon_\theta(\bx_t,t),\quad \bx_T\sim \mathcal{N}(0,\bm I).
    \label{eq:dpmode}
\end{equation}

This probability flows ordinary differential equation (ODE) has the same marginal distribution at each time $t$ as that of the SED in Eq.~\ref{eq:dpmsde}. Sampling can be done by solving the integral of the above ODE from $T$ to $0$. \cite{lu2022dpm} identifies the integral of the above ODE Eq.~\ref{eq:dpmode} has a linear part $f(t)\bx_t$ which can be solved exactly and a nonlinear part $\frac{g^2(t)}{2\sqrt{1-\hat\alpha_t}}\epsilon_\theta(\bx_t,t)$ which needs a black-box ODE solver to approximate. Compared to solving the whole ODE using a black-box solver, this semilinear property enables the elimination of the approximation error of the linear part.
We build our second-order deterministic sampler in a DPM-Solver~\cite{lu2022dpm} way.
To this end, define $\lambda_t=\lambda(t)=\log{\sqrt{(\hat\alpha_t(1-\hat\alpha_t))}}$ and its inverse function $t_\lambda(\cdot)$ such that $t=t_\lambda(\lambda(t))$, we formulate our second-order sampling method on image $\bx_t$ as follows:
\begin{equation}
\begin{aligned}
    s  &= t_\lambda(\frac{\lambda_{t}+\lambda_{t-1}}{2}), \\
    \bm u &= \mathcal{F}_{1st}(f_\theta, \bx_t, s, \bx^C),  \\
    \bx_{t-1} &= \mathcal{F}_{1st}(f_\theta, \bm u, t, \bx^C) . \\
\end{aligned}
\end{equation}
Since there uses first order two times, we call the above iterative sampler a second order deterministic sampler and denote it as $\mathcal{F}_{2ed}(f_\theta, \bx_t, t, \bx^C)$.

\begin{figure*}[!ht]
\centering	
\begin{subfigure}{.3\linewidth}
    \includegraphics [width=1.1\linewidth]{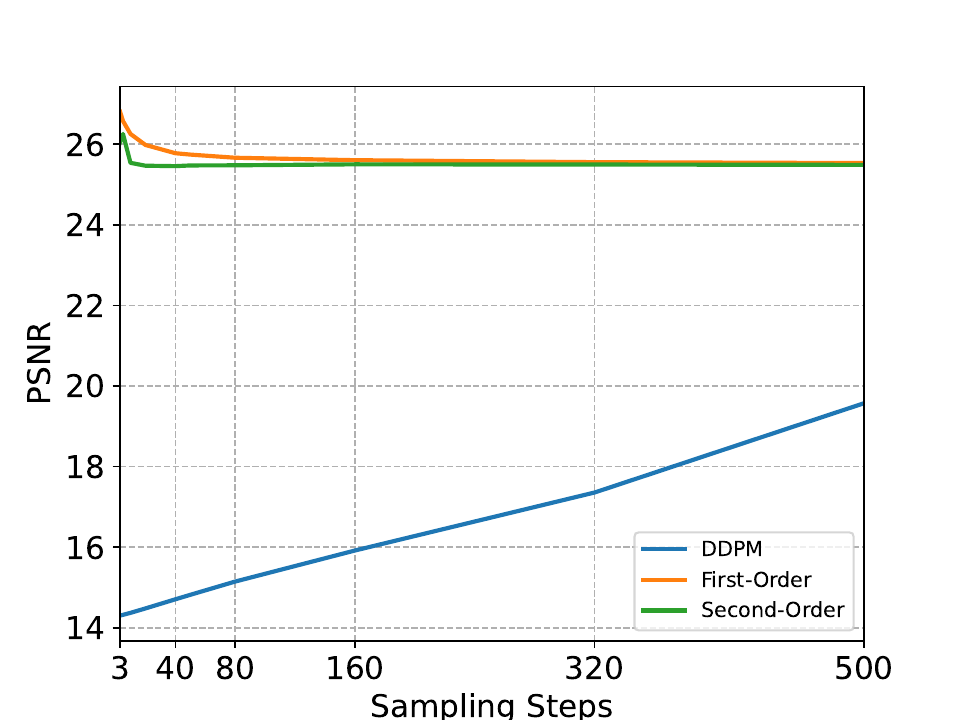}
    \caption{PNSR}
\end{subfigure}
\begin{subfigure}{.3\linewidth}
    \includegraphics [width=1.1\linewidth]{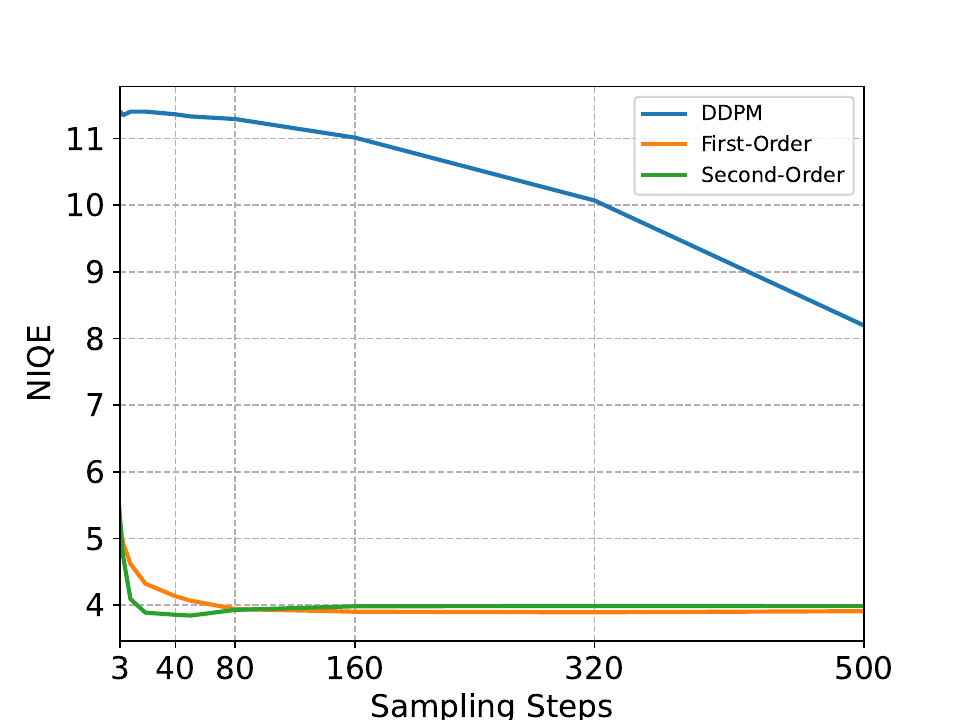}
    \caption{NIQE}
\end{subfigure}
\begin{subfigure}{.3\linewidth}
    \includegraphics [width=1.1\linewidth]{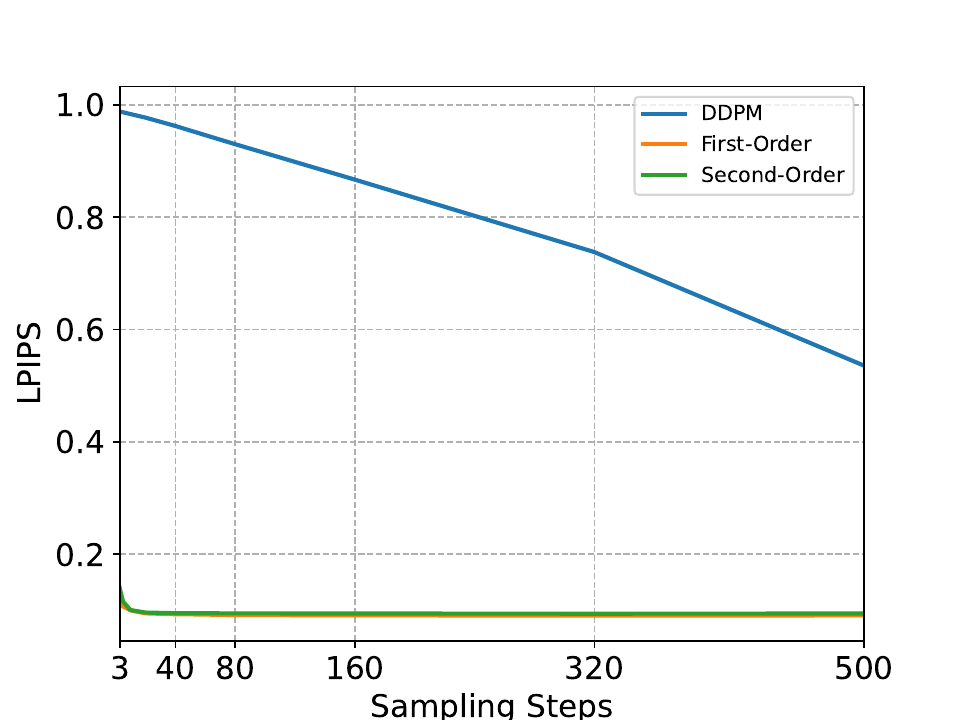}
    \caption{LPIPS}
\end{subfigure}
    \caption{Sampling steps comparison for DDPM sampling method, first-order deterministic, and second-order deterministic sampling method. (Conducted on Urban100 under $\times 4$ scale.)
        }
    \label{fig:sampling_steps_3method}
\end{figure*}

We conducted experiments to compare these three sampling methods.
As shown in Fig.~\ref{fig:sampling_steps_3method}, the PSNR of the original DDPM sampling method is below 20dB in 500 forward steps, which is due to the nature of the stochastic reverse process. DDPM requires many steps to remove the randomness added during each step during the reverse process. First-order and second-order deterministic sampling methods perform much better in small sampling steps. As shown in Fig.~\ref{fig:sampling_steps_3method}, the first-order and second-order methods demonstrate a trade-off between visual quality and image distortion in the low sampling steps region. As the number of sampling steps increases, the PSNR decreases while the NIQE visual quality measure improves. Fig.~\ref{fig:sampling_steps_3method} shows that the second-order sampler can achieve the lowest NIQE scores in just 40 steps. Therefore, we chose second-order sampling as our sampling method and set $T=40$ during inference because we can achieve good perception quality from 40 feedforward steps.

\begin{figure*}[!ht]
\centering	
    \includegraphics [width=1\linewidth]{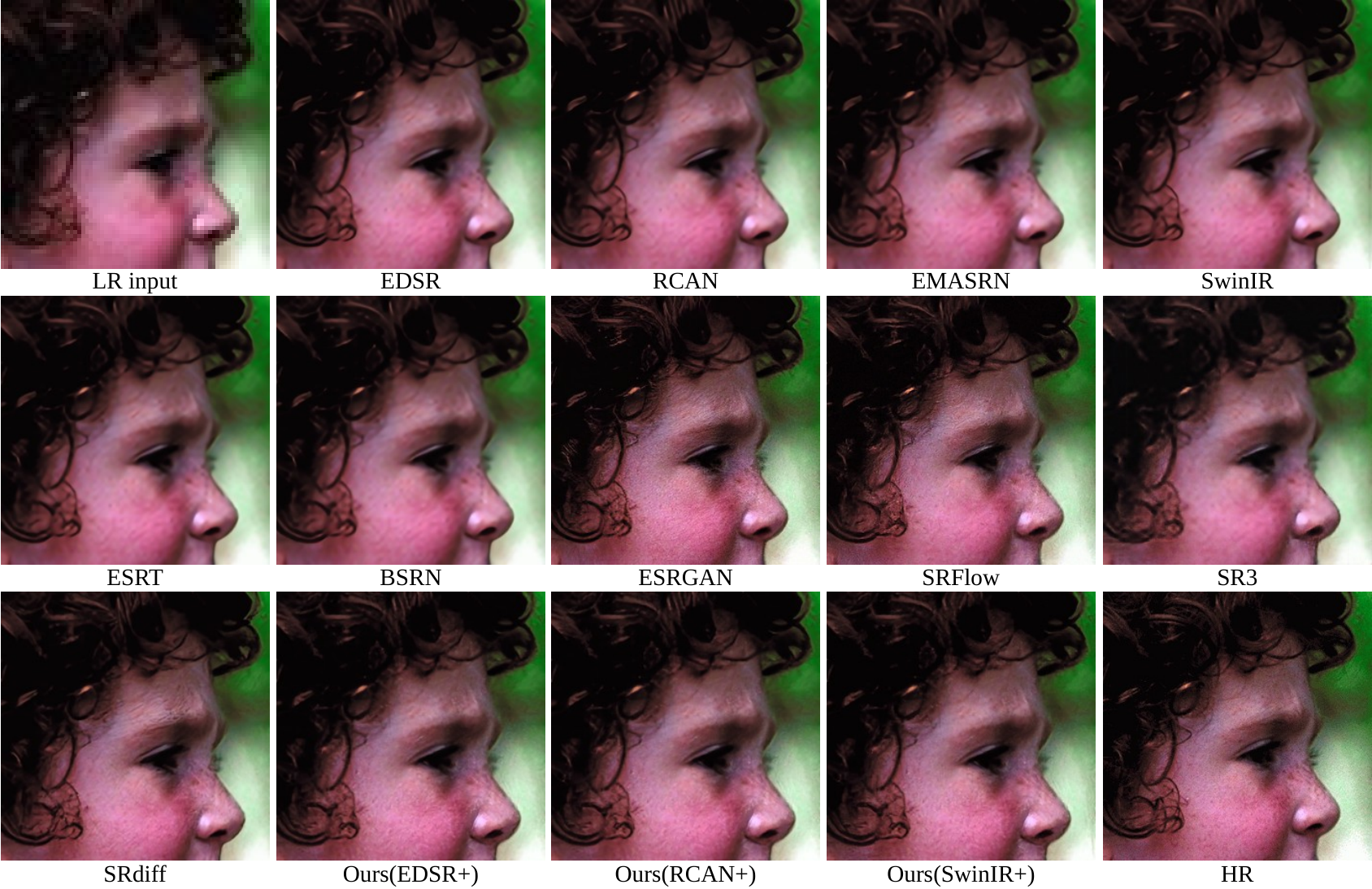}
    \caption{Qualitative comparison with SOTAs performed on image `head' from Set5 ($\times$4 scale, best view in zoomed-in.)}
    \label{fig:main_figure_x4_1}
\end{figure*}

\begin{figure*}[!ht]
\centering	
    \includegraphics [width=1\linewidth]{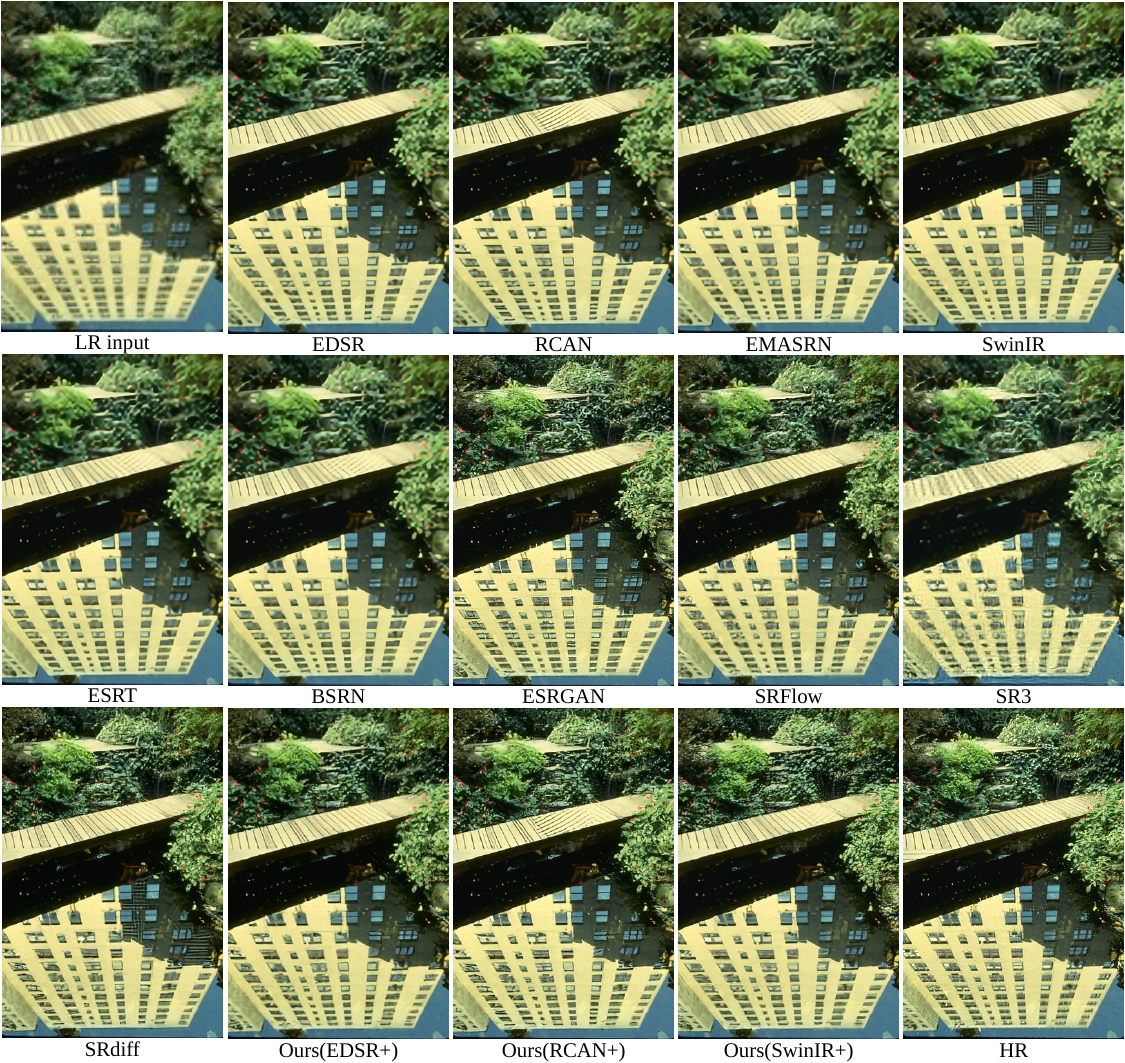}
    \caption{Qualitative comparison with SOTAs performed on image `184026 ' from BSD100 ($\times$4 scale. Cropped and zoomed in for a better view.)}
    \label{fig:main_figure_x4_2}
\end{figure*}

\begin{figure*}[!ht]
\centering	
    \includegraphics [width=1\linewidth]{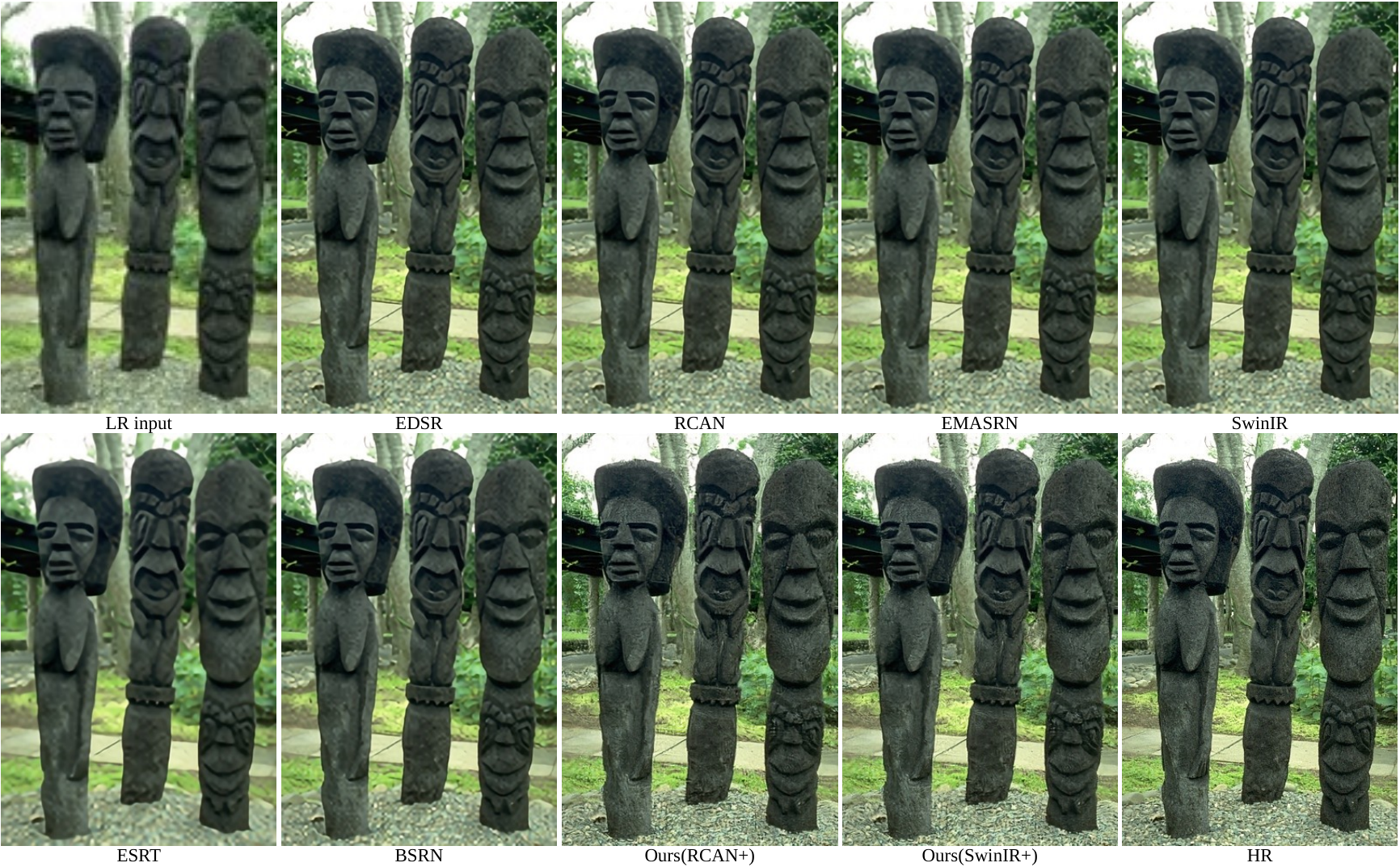}
    \caption{Qualitative comparison with SOTAs performed on image `42012' from BSD100 ($\times$3 scale, best view in zoomed-in.)}
    \label{fig:main_figure_x3_1}
\end{figure*}

\begin{figure*}[!ht]
\centering	
    \includegraphics [width=1\linewidth]{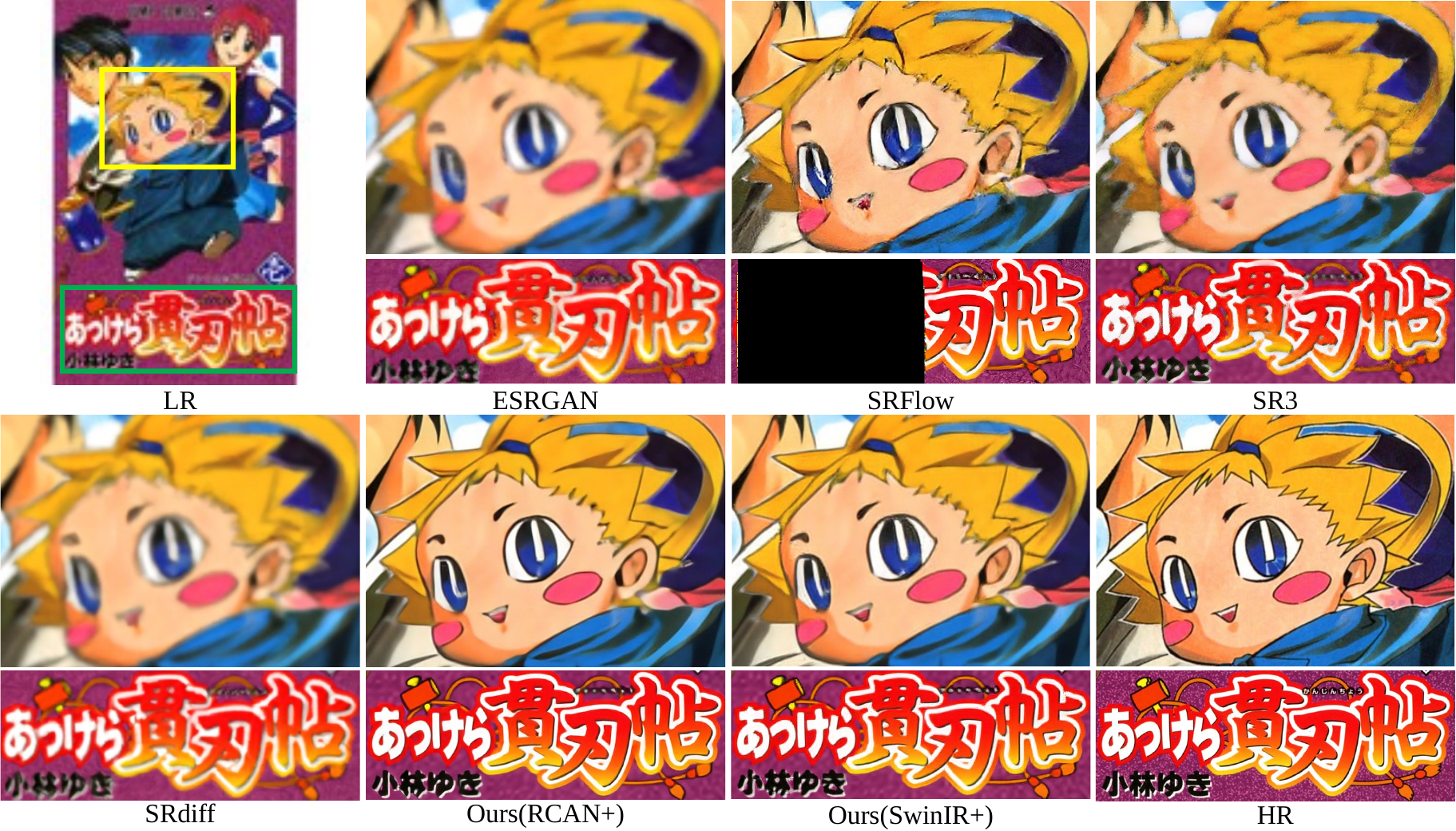}
    \caption{Qualitative comparison with SOTAs performed on image `AkkeraKanjinchou' from Manga109 ($\times$8 scale, best view in zoomed-in.)}
    \label{fig:main_figure_x8_1}
\end{figure*}

\begin{table*}[!ht]
\begin{center}
\caption{Results  on Set5, Set14, BSD100, Urban100, and Manga109. The best and the second-best results are highlighted in \textcolor{red}{red} and \textcolor{green}{green}. } 
\label{tab:results_quantitative_1}
\resizebox{0.8\hsize}{!}{
\begin{tabular}{ccc|c|c|c|c|c|c|c}
\toprule
\hline
\multicolumn{3}{c|}{\multirow{2}{*}{Method} }  & \multirow{2}{*}{ESRGAN}
 & \multirow{2}{*}{SRFlow} & \multirow{2}{*}{SRDiff} & \multirow{2}{*}{SR3} &  \multicolumn{3}{c}{\textbf{Ours}} \\ \cline{8-10}
&&&&&&& \textbf{EDSR+} & \textbf{RCAN+} & \textbf{SwinIR+}\\\midrule
\multicolumn{1}{c|}{\multirow{4}{*}{Set5}}    & \multicolumn{1}{l|}{\multirow{15}{*}{ $\times 4$ }} & LPIPS$\downarrow $  & \textcolor{green}{0.0596} & 0.0767 & 0.0770 & 0.1084 & 0.0619 & 0.0606 &\textcolor{red}{ 0.0564}  \\ \cline{3-10} 
\multicolumn{1}{c|}{}                         &  \multicolumn{1}{l|}{}& PSNR$\uparrow $  &30.459 & 28.35  & 30.938  & 27.314  & 30.830  & \textcolor{green}{30.838}  &\textcolor{red}{31.031} \\ \cline{3-10}
\multicolumn{1}{c|}{}                         & \multicolumn{1}{l|}{} & SSIM$\uparrow $  &0.8516  & 0.8138 &\textcolor{red}{0.8738} & 0.7844 & \textcolor{green}{0.8684} & 0.8645 & 0.8676  \\ \cline{1-1} \cline{3-10}
\multicolumn{1}{c|}{\multirow{4}{*}{Set14}}   & \multicolumn{1}{l|}{}  & LPIPS$\downarrow $ &0.0867 & 0.1318 & 0.1009 & 0.1284 & 0.0883 &\textcolor{green}{0.0865} &\textcolor{red}{0.0827}  \\ \cline{3-10} 
\multicolumn{1}{c|}{}                        &  \multicolumn{1}{l|}{} & PSNR$\uparrow $  &26.282  & 24.97  & \textcolor{red}{27.230}  & 25.475  & 26.996  & 27.015  & \textcolor{green}{27.039}   \\ \cline{3-10} 
\multicolumn{1}{c|}{}                       &  \multicolumn{1}{l|}{}  & SSIM$\uparrow $  & 0.6980  & 0.6908 & \textcolor{red}{0.7432 } & 0.6889 & 0.7316 & 0.7257 & \textcolor{green}{0.7341}  \\ \cline{1-1} \cline{3-10}
\multicolumn{1}{c|}{\multirow{4}{*}{BSD100}} & \multicolumn{1}{l|}{}  & LPIPS$\downarrow $ &\textcolor{red}{0.0834}  & 0.1831 & 0.1041 & 0.1392 &0.0935 &0.0953& \textcolor{red}{0.0834}  \\ \cline{3-10} 
\multicolumn{1}{c|}{}                        &  \multicolumn{1}{l|}{} & PSNR$\uparrow $ &25.288 & 24.654  & \textcolor{red}{25.948}  & 25.208  & 25.743  & 25.865  &\textcolor{green}{25.947}  \\\cline{3-10}
\multicolumn{1}{c|}{}                        &  \multicolumn{1}{l|}{}  & SSIM$\uparrow $ &0.6495 & 0.6573 & \textcolor{red}{0.6833} & 0.6498 & 0.6681 & 0.6706 & \textcolor{green}{0.6743}  \\ \cline{1-1} \cline{3-10}
\multicolumn{1}{c|}{\multirow{4}{*}{Urban100}}& \multicolumn{1}{l|}{}  & LPIPS$\downarrow $ &\textcolor{green}{0.0944}  & 0.1279 & 0.1077 & 0.1993 &0.0997 &0.0997 & \textcolor{red}{0.0934}  \\ \cline{3-10} 
\multicolumn{1}{c|}{}    & \multicolumn{1}{l|}{} & PSNR$\uparrow $  &24.349  & 23.652  & 25.340  & 22.489  & 25.452  & \textcolor{green}{25.587}  &\textcolor{red}{25.852}   \\ \cline{3-10}
\multicolumn{1}{c|}{}                         & \multicolumn{1}{l|}{}  & SSIM$\uparrow $ & 0.7327 & 0.7312 & 0.7661 & 0.6336 & 0.7649 &\textcolor{green}{ 0.7681} &\textcolor{red}{0.7796}  \\ \cline{1-1} \cline{3-10}
\multicolumn{1}{c|}{\multirow{4}{*}{Manga109}}& \multicolumn{1}{l|}{}  & LPIPS$\downarrow $ &0.0420  & 0.0660 & 0.0473 & 0.1100 &0.0409   &\textcolor{green}{0.0396} &\textcolor{red}{0.0374}  \\ \cline{3-10} 

\multicolumn{1}{c|}{}                       &  \multicolumn{1}{l|}{}  & PSNR$\uparrow $ &28.476 & 27.14  & 28.668  & 24.691  &29.072 & \textcolor{green}{29.385}  &\textcolor{red}{29.601}  \\ \cline{3-10} 
\multicolumn{1}{c|}{}                        &  \multicolumn{1}{l|}{} & SSIM$\uparrow $ &0.8595  & 0.8244 & \textcolor{green}{0.8851} & 0.7568 & 0.8791 & 0.8816 &\textcolor{red}{0.8874}  \\  \midrule

\multicolumn{1}{c|}{\multirow{4}{*}{Set5}} &\multicolumn{1}{l|}{\multirow{15}{*}{ $\times 8$ }}     & LPIPS$\downarrow $ &0.2626&0.2304 &0.3129 & 0.1872& 0.2988 &\textcolor{green}{0.1813}  & \textcolor{red}{0.1669} \\ \cline{3-10} 
\multicolumn{1}{c|}{}                      & \multicolumn{1}{l|}{}  & PSNR$\uparrow $  & 24.830  &22.604 &20.074 &25.394  &24.955 &\textcolor{green}{27.026}  &\textcolor{red}{27.165}  \\ \cline{3-10} 
\multicolumn{1}{c|}{}                      & \multicolumn{1}{l|}{}    & SSIM$\uparrow $  & 0.6843  &0.6062 &0.5852 &0.6897 &0.6897  &\textcolor{green}{0.7834} &\textcolor{red}{0.7842} \\ \cline{1-1}\cline{3-10}
\multicolumn{1}{c|}{\multirow{4}{*}{Set14}} &  \multicolumn{1}{l|}{}   & LPIPS$\downarrow $& 0.2536 &0.2965 &0.2929 & \textcolor{red}{0.1983}& 0.2740  &\textcolor{green}{0.2083}  &0.2108  \\ \cline{3-10}
\multicolumn{1}{c|}{}                      &  \multicolumn{1}{l|}{}   & PSNR$\uparrow $   & 23.496 &21.261 &21.149 &23.959 &23.648 &\textcolor{red}{25.136}  & \textcolor{green}{25.133}  \\  \cline{3-10}
\multicolumn{1}{c|}{}                      &  \multicolumn{1}{l|}{}   & SSIM$\uparrow $   & 0.5854 &0.4894 &0.5393 &0.5850 &0.5864 &\textcolor{red}{0.6487}   &\textcolor{green}{0.6485}  \\ \cline{1-1} \cline{3-10}
\multicolumn{1}{c|}{\multirow{4}{*}{BSD100}}& \multicolumn{1}{l|}{}   & LPIPS$\downarrow $ & 0.2503&0.3236 &0.2879 &\textcolor{red}{0.1975} &0.2734 &0.2202   &\textcolor{green}{0.2219}	 \\ \cline{3-10} 
\multicolumn{1}{c|}{}                       &  \multicolumn{1}{l|}{}  & PSNR$\uparrow $   & 23.868 &21.619 &19.162 & 23.918&24.015 &\textcolor{red}{24.983} &\textcolor{green}{24.944}   \\ \cline{3-10} 
\multicolumn{1}{c|}{}                       &  \multicolumn{1}{l|}{}  & SSIM$\uparrow $   &0.5518  &0.4634 &0.4560 & 0.5345&0.5572 & \textcolor{red}{0.6054}  & \textcolor{green}{0.6052} \\ \cline{1-1}   \cline{3-10}
\multicolumn{1}{c|}{\multirow{4}{*}{Urban100}}& \multicolumn{1}{l|}{} & LPIPS$\downarrow $ &0.3062 &0.2968 &0.3494 & 0.2722   & 0.3384   & \textcolor{red}{0.2177} & \textcolor{green}{0.2245} \\   \cline{3-10} 
\multicolumn{1}{c|}{}                        &  \multicolumn{1}{l|}{} & PSNR$\uparrow $   &20.977  &19.383 &19.659 &21.538 &21.164 &\textcolor{red}{22.947}   &\textcolor{green}{22.932}  \\   \cline{3-10} 
\multicolumn{1}{c|}{}                        &  \multicolumn{1}{l|}{} & SSIM$\uparrow $   & 0.5359 &0.4999 &0.5155 &0.5546 &0.5390 & \textcolor{red}{0.6428}  &\textcolor{green}{0.6413} \\ \cline{1-1}  \cline{3-10}
\multicolumn{1}{c|}{\multirow{4}{*}{Manga109}}& \multicolumn{1}{l|}{} & LPIPS$\downarrow $ &0.2364 &0.2185 &0.2660 &0.1820 &0.2564 &\textcolor{red}{0.1239}   & \textcolor{green}{0.1322}\\ %\cline{3-9} 
\multicolumn{1}{c|}{}                         & \multicolumn{1}{l|}{} & PSNR$\uparrow $   &22.166 &20.480 &18.757 &23.003 &22.145 &\textcolor{red}{25.097}  & \textcolor{green}{25.019} \\  \cline{3-10}
\multicolumn{1}{c|}{}                         & \multicolumn{1}{l|}{} & SSIM$\uparrow $  &0.6858   &0.6434 &0.6311 &0.7126 &0.6790 &\textcolor{red}{0.7984} &\textcolor{green}{0.7931} \\
\hline
\bottomrule
\end{tabular}}
\end{center}
\end{table*}

\begin{table*}[!ht]
\caption{Results of different scales on Set5, Set14, BSD100, Urban100, and Manga109. The \textbf{bold} represents the best result. } 
\label{tab:results_quantitative_2}
\centering
\resizebox{0.9\hsize}{!}{
% \begin{center}
\begin{tabular}{ccc|c|c|c|c|c| c| c|c|c}
\toprule
\hline

\multicolumn{3}{c|}{\multirow{2}{*}{Method} }
& \multirow{2}{*}{EDSR}   & \multirow{2}{*}{RCAN}    & \multirow{2}{*}{EMASRN } &\multirow{2}{*}{SwinIR}  & \multirow{2}{*}{ESRT}  & \multirow{2}{*}{BSRN} &  \multicolumn{3}{c}{\textbf{Ours}} \\ \cline{10-12}
&&&&&&&&& \textbf{EDSR+} & \textbf{RCAN+} & \textbf{SwinIR+}\\\midrule
\multicolumn{1}{c|}{\multirow{4}{*}{Set5}}  &\multicolumn{1}{l|}{\multirow{20}{*}{ $\times 2$ }}  & LPIPS$\downarrow $  &0.0322 &0.0321   &-   &0.0316 &0.0609 &0.0611  &\textbf{0.0121} &\textbf{0.0121}  &0.0125  \\ \cline{3-12} 
\multicolumn{1}{c|}{}                       & \multicolumn{1}{l|}{}  & NIQE$\downarrow $   &5.3005 &5.2721  &-  &5.3325 & 5.3226 &5.3824  &4.3374 &\textbf{4.4035}  &4.4151  \\ \cline{3-12}  
\multicolumn{1}{c|}{}                       & \multicolumn{1}{l|}{}   & PSNR$\uparrow $     &38.193 &38.271   & - &\textbf{38.357} &38.088 &38.072 &36.241 &36.255  &36.462  \\ \cline{3-12} 
\multicolumn{1}{c|}{}                       & \multicolumn{1}{l|}{}   & SSIM$\uparrow $     &0.9609 &0.9614  &-  &\textbf{0.9620}   &0.9598  &0.9597  &0.9404  &0.9399  &0.9431 \\   \cline{1-1} \cline{3-12}
\multicolumn{1}{c|}{\multirow{4}{*}{Set14}}  & \multicolumn{1}{l|}{}   & LPIPS$\downarrow $ &0.0458 &0.0446    & -   &0.0433 &0.0968 &0.0937   &0.0275  &0.0262  & \textbf{0.0252} \\ \cline{3-12} 
\multicolumn{1}{c|}{}                       & \multicolumn{1}{l|}{}   & NIQE$\downarrow $   &5.0109 &4.9893   &-  &4.9729  &5.2071 &5.1882  &\textbf{3.8002} &3.8680  &3.8216  \\ \cline{3-12} 
\multicolumn{1}{c|}{}                       & \multicolumn{1}{l|}{}   & PSNR$\uparrow $     &33.948 &34.126    &-   &\textbf{34.141}   &33.690 & 33.642   &32.073 &32.315  & 32.277  \\  \cline{3-12} 
\multicolumn{1}{c|}{}                       & \multicolumn{1}{l|}{}   & SSIM$\uparrow $     &0.9202 &0.9216    &-  &\textbf{0.9227}  &0.9183 &0.9186  &0.8834 & 0.8858 &0.8863  \\ \cline{1-1} \cline{3-12}
\multicolumn{1}{c|}{\multirow{4}{*}{BSD100}}& \multicolumn{1}{l|}{}   & LPIPS$\downarrow $  &0.0623 &0.0615   &-   &0.0608 &0.1463 &0.1458   &0.0281 &\textbf{0.0278}  &0.0283 \\ \cline{3-12} 
\multicolumn{1}{c|}{}                       & \multicolumn{1}{l|}{}   & NIQE$\downarrow $   &4.9536 & 4.9673   & -   &4.9238 &5.1657 & 5.1902 &\textbf{3.3873} &3.5461  &3.4376  \\ \cline{3-12} 
\multicolumn{1}{c|}{}                       & \multicolumn{1}{l|}{}   & PSNR$\uparrow $     &32.352 &32.389    & -  &\textbf{32.448} &32.272 &32.221   &30.176 &30.293  &30.346   \\ \cline{3-12} 
\multicolumn{1}{c|}{}                       & \multicolumn{1}{l|}{}   & SSIM$\uparrow $     &0.9019 &0.9024   & -  &\textbf{0.9030} &0.8993 &0.8987  & 0.8544 & 0.8572 &0.8588  \\ \cline{1-1} \cline{3-12}
\multicolumn{1}{c|}{\multirow{4}{*}{Urban100}}&\multicolumn{1}{l|}{}  & LPIPS$\downarrow $  &0.0359 &0.0346    &-& 0.0333  & 0.0619&0.0619   &0.0273 & \textbf{0.0269}  &0.0272  \\ \cline{3-12} 
\multicolumn{1}{c|}{}                        & \multicolumn{1}{l|}{}  & NIQE$\downarrow $   &4.5070 &4.4983   &-   &4.4880 &4.6086 &4.5908  &3.9641 &3.9850  &\textbf{3.9371} \\ \cline{3-12} 
\multicolumn{1}{c|}{}                        &\multicolumn{1}{l|}{}   & PSNR$\uparrow $     &32.967 &33.175   &-   &\textbf{33.404}  &32.602 &32.324&31.386 &31.538  &31.721  \\ \cline{3-12} 
\multicolumn{1}{c|}{}                        & \multicolumn{1}{l|}{}  & SSIM$\uparrow $     &0.9359 &0.9371    &-  &\textbf{0.9394} &0.9320 & 0.9296 &0.9124 &  0.9112 &0.9152 \\ \cline{1-1} \cline{3-12}
\multicolumn{1}{c|}{\multirow{4}{*}{Manga109}}&\multicolumn{1}{l|}{}  & LPIPS$\downarrow $  &0.0106 &0.0102  &-  &0.0100  & 0.0228&0.0226    &0.0072 &0.0070  & \textbf{0.0067}\\ 	\cline{3-12} 
\multicolumn{1}{c|}{}                         &\multicolumn{1}{l|}{}  & NIQE$\downarrow $   &4.5104 &4.5217  &-  &4.4956  & 4.6483&4.6622   &\textbf{3.8334} &3.9864  &3.8553 \\ \cline{3-12} 
\multicolumn{1}{c|}{}                         &\multicolumn{1}{l|}{}  & PSNR$\uparrow $     &39.193 &39.438  &-  &\textbf{39.586}  & 39.073&38.992   &37.046 &37.518  &37.607  \\ \cline{3-12} 
\multicolumn{1}{c|}{}                        & \multicolumn{1}{l|}{}  & SSIM$\uparrow $     &0.9782 &0.9787   & -  &\textbf{0.9791}  &0.9773 &0.9771 &0.9650 &0.9660  &0.9669 \\ \midrule
\multicolumn{1}{c|}{\multirow{4}{*}{Set5}}   & \multicolumn{1}{l|}{\multirow{20}{*}{ $\times 3$ }}  & LPIPS$\downarrow $  &0.0758 &0.0747   &0.1356   &0.0734 &0.1363 & 0.1378  &0.0365 &\textbf{0.0354}  &0.0363 \\ \cline{3-12} 
\multicolumn{1}{c|}{}                        &\multicolumn{1}{l|}{}   & NIQE$\downarrow $   &6.4616 &6.4571   & 6.5556  &6.6240 & 6.6755&6.8924  &5.0188 &\textbf{4.8185}  &4.8930  \\ \cline{3-12} 
\multicolumn{1}{c|}{}                        &\multicolumn{1}{l|}{}   & PSNR$\uparrow $     &34.680 &34.758    & 34.361   &\textbf{34.878} &34.612 &34.499  &32.618 &32.715  &33.001  \\ \cline{3-12} 
\multicolumn{1}{c|}{}                        &\multicolumn{1}{l|}{}   & SSIM$\uparrow $     &0.9294 &0.9300   &0.9264  &\textbf{0.9312} &0.9271 & 0.9262 &0.8989 & 0.9002  &0.9059 \\ \cline{1-1} \cline{3-12}
\multicolumn{1}{c|}{\multirow{4}{*}{Set14}}   & \multicolumn{1}{l|}{}  & LPIPS$\downarrow $ &0.1002 &0.1001   & 0.2175  &0.0976 &0.2288 &0.2092 &0.0630 & \textbf{0.0607} &0.0637  \\ \cline{3-12} 
\multicolumn{1}{c|}{}                        & \multicolumn{1}{l|}{}  & NIQE$\downarrow $   &5.5798 &5.6797    & 5.9351  &5.6477  &5.9953 & 5.9011  &3.8107 &3.8364  &\textbf{ 3.7388}  \\ \cline{3-12} 
\multicolumn{1}{c|}{}                        & \multicolumn{1}{l|}{}  & PSNR$\uparrow $     &30.533 &30.627   &28.571  &\textbf{30.771}  &30.583 & 30.379 &28.423 &28.578  &28.762   \\ \cline{3-12} 
\multicolumn{1}{c|}{}                        &\multicolumn{1}{l|}{}   & SSIM$\uparrow $     &0.8465 &0.8476  &0.7809   &\textbf{0.8502} &0.8341 & 0.8435 &0.7861 &0.7898  &0.7953  \\ \cline{1-1} \cline{3-12}
\multicolumn{1}{c|}{\multirow{4}{*}{BSD100}} &\multicolumn{1}{l|}{}   & LPIPS$\downarrow $  &0.1163 &0.1150   & 0.2967 &0.1124  &0.2968 & 0.2944  &0.0641 &0.0648  &\textbf{0.0634} \\ \cline{3-12} 
\multicolumn{1}{c|}{}                       & \multicolumn{1}{l|}{}   & NIQE$\downarrow $   &5.7653 &5.8292   &6.0468   &5.7018 &6.2079 & 6.0124  &3.4016 & 3.4324 &\textbf{3.3014}  \\ \cline{3-12} 
\multicolumn{1}{c|}{}                       & \multicolumn{1}{l|}{}   & PSNR$\uparrow $     &29.263 &29.301   &29.053  &\textbf{29.367} &29.224 &29.181&26.938 & 27.139 & 27.163  \\ \cline{3-12} 
\multicolumn{1}{c|}{}                       & \multicolumn{1}{l|}{}   & SSIM$\uparrow $     &0.8096 &0.8106    &0.8035  &\textbf{0.8124} &0.8049 &0.8035   &0.7355 &0.7415  &0.7417  \\ \cline{1-1} \cline{3-12}
\multicolumn{1}{c|}{\multirow{4}{*}{Urban100}}&\multicolumn{1}{l|}{}  & LPIPS$\downarrow $  &0.0863 &0.0830  & 0.1675&0.0798  &0.1674 &0.1581  &0.0661 & 0.0654 &\textbf{0.0647}  \\ \cline{3-12} 
\multicolumn{1}{c|}{}                         & \multicolumn{1}{l|}{} & NIQE$\downarrow $   &5.0547 &5.1298 & 5.2835 &5.0891  &5.3741 &5.2855  &4.0667 &4.0781  & \textbf{4.0287}\\ \cline{3-12} 
\multicolumn{1}{c|}{}                         &\multicolumn{1}{l|}{}  & PSNR$\uparrow $     &28.812 &29.009   &28.042  &\textbf{29.288} &28.469 & 28.389  &27.424 & 27.722 &27.889  \\ \cline{3-12} 
\multicolumn{1}{c|}{}                         &\multicolumn{1}{l|}{}  & SSIM$\uparrow $     &0.8659 &0.8685  &0.8493  &\textbf{0.8744}  & 0.8578&0.8558   &0.8309 &0.8361  &0.8412 \\ \cline{1-1} \cline{3-12}
\multicolumn{1}{c|}{\multirow{4}{*}{Manga109}} &\multicolumn{1}{l|}{} & LPIPS$\downarrow $  &0.0328 &0.0320  & 0.0662  &0.0307  &0.0669 & 0.0638  &0.0231 & 0.0215 & \textbf{0.0220} \\ \cline{3-12} 
\multicolumn{1}{c|}{}                         &\multicolumn{1}{l|}{}  & NIQE$\downarrow $   &4.8532 &4.9141   & 4.9435 &4.8789 & 5.0512& 4.9802  &3.9195 &3.8586  &\textbf{3.6899} \\ \cline{3-12} 
\multicolumn{1}{c|}{}                        & \multicolumn{1}{l|}{}  & PSNR$\uparrow $     &34.200 &34.429  &33.433   &\textbf{34.749}  &34.109 &33.982   &32.207 &32.338  &32.428  \\ \cline{3-12} 
\multicolumn{1}{c|}{}                        & \multicolumn{1}{l|}{}  & SSIM$\uparrow $     &0.9486 &0.9498   &0.9433   &\textbf{0.9517}  &0.9454 &0.9450 &0.9245 &0.9236  &0.9259 \\ 
\midrule
\multicolumn{1}{c|}{\multirow{4}{*}{Set5}}   &\multicolumn{1}{l|}{\multirow{20}{*}{ $\times 4$ }}  & LPIPS$\downarrow $ & 0.1098 & 0.1096  & 0.1820 & 0.1087  &0.1889 & 0.1865 & 0.0619 & 0.0606  &\textbf{0.0564}  \\ \cline{3-12} 
\multicolumn{1}{c|}{}                        &\multicolumn{1}{l|}{}  & NIQE$\downarrow $  & 7.2500 & 7.1562  & 7.2289  &7.0368 &6.9859 & 7.2315  & 5.5288 & 5.6325 &\textbf{4.9999}  \\ \cline{3-12} 
\multicolumn{1}{c|}{}                        & \multicolumn{1}{l|}{} & PSNR$\uparrow $  & 32.426  & 32.638   &32.173   & \textbf{32.722} & 32.442&32.387  & 30.830  & 30.838  &31.031 \\ \cline{3-12} 
\multicolumn{1}{c|}{}                       & \multicolumn{1}{l|}{}  & SSIM$\uparrow $  & 0.8985 & 0.9002  &0.8948   & \textbf{0.9021} &0.8960 & 0.8949  & 0.8684 & 0.8645 & 0.8676  \\ \cline{1-1} \cline{3-12}
\multicolumn{1}{c|}{\multirow{4}{*}{Set14}}   & \multicolumn{1}{l|}{} & LPIPS$\downarrow $ & 0.1415 & 0.1387 &0.2886  &0.1369  &0.2911   & 0.2871  & 0.0883 &0.0865 &\textbf{0.0827}  \\ \cline{3-12} 
\multicolumn{1}{c|}{}                       & \multicolumn{1}{l|}{}  & NIQE$\downarrow $  & 6.0475 & 6.1797   &6.3646  & 6.2370 &6.3369 &6.2940  &3.8035 & 3.8188 & \textbf{3.7958}  \\ \cline{3-12} 
\multicolumn{1}{c|}{}                        &\multicolumn{1}{l|}{}  & PSNR$\uparrow $  & 28.679  & 28.851   & 28.572   &\textbf{28.937} &28.614 & 28.534 & 26.996  & 27.015  & 27.039   \\ \cline{3-12} 
\multicolumn{1}{c|}{}                        & \multicolumn{1}{l|}{} & SSIM$\uparrow $  & 0.7883 & 0.7885   &0.7809  &  \textbf{0.7914} &0.7845 & 0.7837 & 0.7316 & 0.7257 & 0.7341  \\ \cline{1-1} \cline{3-12}
\multicolumn{1}{c|}{\multirow{4}{*}{BSD100}} &\multicolumn{1}{l|}{}  & LPIPS$\downarrow $ & 0.1551 & 0.1536   & 0.3847  & 0.1542 &0.3881 &0.3829   &0.0935 &0.0953 & \textbf{0.0834}  \\ \cline{3-12} 
\multicolumn{1}{c|}{}                        & \multicolumn{1}{l|}{} & NIQE$\downarrow $  & 6.3351 & 6.3104 & 6.5912 & 6.3638 &6.6465 &6.5235   &3.3751 & 3.4251 &\textbf{3.4096}  \\ \cline{3-12} 
\multicolumn{1}{c|}{}                        &\multicolumn{1}{l|}{}  & PSNR$\uparrow $  & 27.734  & 27.743    &27.552    & \textbf{27.841} &27.725 &27.675  & 25.743  & 25.865  &25.947   \\ \cline{3-12} 
\multicolumn{1}{c|}{}                       & \multicolumn{1}{l|}{}  & SSIM$\uparrow $  & 0.7425 & 0.7430  &0.7351    & \textbf{0.7461} &0.7369 &0.7353  & 0.6681 & 0.6706 & 0.6743  \\ \cline{1-1} \cline{3-12}
\multicolumn{1}{c|}{\multirow{4}{*}{Urban100}}& \multicolumn{1}{l|}{} & LPIPS$\downarrow $ & 0.1220 & 0.1220  &0.2342  & 0.1200  &0.2396 & 0.2315 &0.0997 &0.0997 & \textbf{0.0934}  \\ \cline{3-12} 
\multicolumn{1}{c|}{}                        & \multicolumn{1}{l|}{} & NIQE$\downarrow $  & 5.4302 & 5.4886  &5.6733   & 5.4203   &5.9356 &5.7585   &4.0943 &\textbf{4.0521} & 4.1578  \\   \cline{3-12} 
\multicolumn{1}{c|}{}                       &\multicolumn{1}{l|}{}   & PSNR$\uparrow $  & 26.645  & 26.745  &26.012  & \textbf{27.075} &26.522 & 26.278 & 25.452  &25.587  &25.852   \\ \cline{3-12} 
\multicolumn{1}{c|}{}                        &\multicolumn{1}{l|}{}  & SSIM$\uparrow $  & 0.8039 & 0.8066  & 0.7837   & \textbf{0.8165} &0.7965 &0.7903 & 0.7649 & 0.7681 & 0.7796  \\ \cline{1-1} \cline{3-12}
\multicolumn{1}{c|}{\multirow{4}{*}{Manga109}}&\multicolumn{1}{l|}{} & LPIPS$\downarrow $ & 0.0562 & 0.0544 &0.1066  & 0.1033  &0.1109 & 0.1035  &0.0409   &0.0396 & \textbf{0.0374}  \\ \cline{3-12} 
\multicolumn{1}{c|}{}                        & \multicolumn{1}{l|}{} & NIQE$\downarrow $  & 5.1480  & 5.2272  & 5.2393 & 5.1456 &5.4343 &5.3175    & 3.7661  & \textbf{3.7599} & 3.7985  \\ \cline{3-12} 
\multicolumn{1}{c|}{}                        & \multicolumn{1}{l|}{} & PSNR$\uparrow $  & 31.057 &31.197  &30.413  & \textbf{31.668} & 30.979&30.837   &29.072 & 29.385  & 29.601  \\ \cline{3-12} 
\multicolumn{1}{c|}{}                         &\multicolumn{1}{l|}{} & SSIM$\uparrow$  & 0.9160 & 0.9170  &0.9076    & \textbf{0.9226}  &0.9107 & 0.9097 & 0.8791 & 0.8816 & 0.8874  \\ 
\hline
\bottomrule
\end{tabular}
}
\end{table*}

\section{Experiments}

\subsection{Experimental Settings}
% \noindent
\textbf{Dataset.} We use 800 image pairs in DIV2K as the training set. We take public benchmark datasets,~\ie, Set5, Set14, Urnban100, BSD100, and Manga109 as the test set to compare with other methods.

\textbf{Setups.} We set $T=1000$ for training and $T=40$ during the inference time for the diffusion model. We take the pre-trained super-resolution models ( EDSR~\cite{lim2017enhanced}, and RCAN~\cite{zhang2018image}, SwinIR~\cite{liang2021swinir}) to provide the initial super-resolution image,~\ie the conditional image. The conditional diffusion model is trained with Adam optimizer and batch size 16, with a learning rate of $1\times 10^{-4}$ for 400k steps. The architecture of the model is the same as that in~\cite{saharia2022image}. 

\textbf{Metrics.} The previous study has shown that distortion and perceptual quality are at odds with each other, and there is a trade-off between them~\cite{blau2018perception}. Since our work focuses on the perceptual quality, except the distortion metrics: PSNR and SSIM, we also provide perceptual metrics:  LPIPS~\cite{zhang2018unreasonable} and NIQE~\cite{mittal2012making}
to show that our method can generate better perceptual results than other methods. LPIPS is recently introduced as a reference-based image quality evaluation metric, which computes the perceptual similarity between the ground truth and the SR image. NIQE is a no-reference image quality score built on a “quality aware” collection of statistical features based on a simple and successful space domain natural scene statistic model.

\subsection{Quantitative and Qualitative Results}

To verify the effectiveness of our ACDMSR, we select some SOTA generative methods to conduct the comparative experiments, including  ESRGAN~\cite{wang2018esrgan}, SRFlow~\cite{lugmayr2020srflow}, SRDiff~\cite{li2022srdiff}, SR3~\cite{saharia2022image}. We selected EDSR~\cite{lim2017enhanced}, RCAN~\cite{zhang2018image}, and SwinIR~\cite{liang2021swinir} to provide the conditional image, respectively. Therefore, we report three cases for our cDPMASR,~\ie, ~\textbf{EDSR+},~\textbf{RCAN+}, and~\textbf{SwinIR+}. 
In addition, we also compare our method with some SOTA tradition CNN-based SR methods to verify further the effectiveness of our ACDMSR, including  EDSR~\cite{lim2017enhanced}, RCAN~\cite{zhang2018image}, EMASRN~\cite{zhu2021lightweight}, SwinIR~\cite{liang2021swinir}, ESRT~\cite{lu2022transformer}, and BSRN~\cite{li2022blueprint}. All the results are obtained from the provided codes or publicized papers. 

Tab.~\ref{tab:results_quantitative_1} reports the PSNR, SSIM, and LPIPS values for those generative methods. Our method achieves superior performance under these quantitative metrics in terms of both distortion and perceptual quality across multiple standard datasets. ESRGAN is a typical GAN-based SR method, which includes an SR image generator and an SR image discriminator to push the generator to generate more realistic images. It achieves better LPIPSs,  lower PSNRs, and lower SSIMs on different datasets under different scales compared with SRDif. It seems the results generated by ESRGAN in Fig.~\ref{fig:main_figure_x4_1}, Fig.~\ref{fig:main_figure_x4_2} and Fig.\ref{fig:main_figure_x8_1} include more details than other methods, but it introduces too many false artifacts compared to the ground truth. SRFlow adopts the flow model to obtain reasonable high-resolution images by learning a conditional distribution when given low-resolution images. But the flow model needs invertible parameterized transformations with a tractable Jacobian determinant, which limits their expressiveness~\cite{saharia2022image} and obtains worse LPIPSs, lower PSNRs, and lower SSIM compared with SRdiff and our method. And the results of SRflow seem noisy. To our knowledge, SRDiff and SR3 are state-of-the-art SR methods based on the diffusion model. SRDiff employs a two-stage structure, first pre-training an SR model and then optimizing the diffusion model. SR3 proposes an intuitive SR diffusion model based on the standard diffusion model in~\cite{ho2020denoising}. Our method is similar to these two methods. However, we use existing SR methods to provide the conditional image instead of pretraining a new conditional-provided model and adjusting the optimization method by predicting the original image instead of the noise, which is more suitable for the SR task. With better conditional image, our method exhibits superior performance on both quantitative and qualitative results than SR3~\cite{saharia2022image}. Though SRdiff obtains some comparable numeric results in Tab.\ref{tab:results_quantitative_1}, the visual results of our ACDMSR are closer to ground truths (Especially the forehead in Fig.\ref{fig:main_figure_x4_1}, the plants and the building in Fig.\ref{fig:main_figure_x4_2}). In Sec.\ref{sec:ab}, we have further conducted ablation studies to prove that a better conditional image indeed helps improve the SR performance of the diffusion model.
% and sampling process of  the standard diffusion model

Tab.~\ref{tab:results_quantitative_2} reports the PSNR, SSIM, LPIPS, and NIQE values for those traditional CNN-based SR methods. Because these methods are PSNR-directed and they all focus on obtaining results with good distortion~\cite{blau2018perception}, they can perform well on PSNR and SSIM, which are well-known to only partially correspond to human perception and can lead to algorithms with visibly lower quality in the reconstructed images~\cite{saharia2022image}. The SR results of these PSNR-oriented methods are obviously so over-smooth that some details are missing. Though the PSNR and SSIM numbers of our method are slightly lower than theirs, it performs better when considering the metrics more in line with the human visual system. 

In addition, we present Fig.~\ref{fig:main_figure_x4_1}, Fig.~\ref{fig:main_figure_x4_2}, Fig.~\ref{fig:main_figure_x3_1}, and Fig.~\ref{fig:main_figure_x8_1} to illustrate the SR visual results on different datasets with varying scales. Our methods perform well on a variety of content, including humans, plants, text, and animals. These results further demonstrate the effectiveness of our approach in achieving both metric and perceptual quality.

\subsection{Ablation Study}
\label{sec:ab}
In this section, we conduct ablation studies to verify the influence of different conditional images on our ACDMSR. In addition, we also investigate how stochastic sampling and deterministic sampling influence the reconstruction results. Furthermore, we conduct experiments to verify the effectiveness of different loss functions.

\label{ab:different_conditional_inputs}
\begin{figure*}[!ht]
\centering	
    \includegraphics [width=1\linewidth]{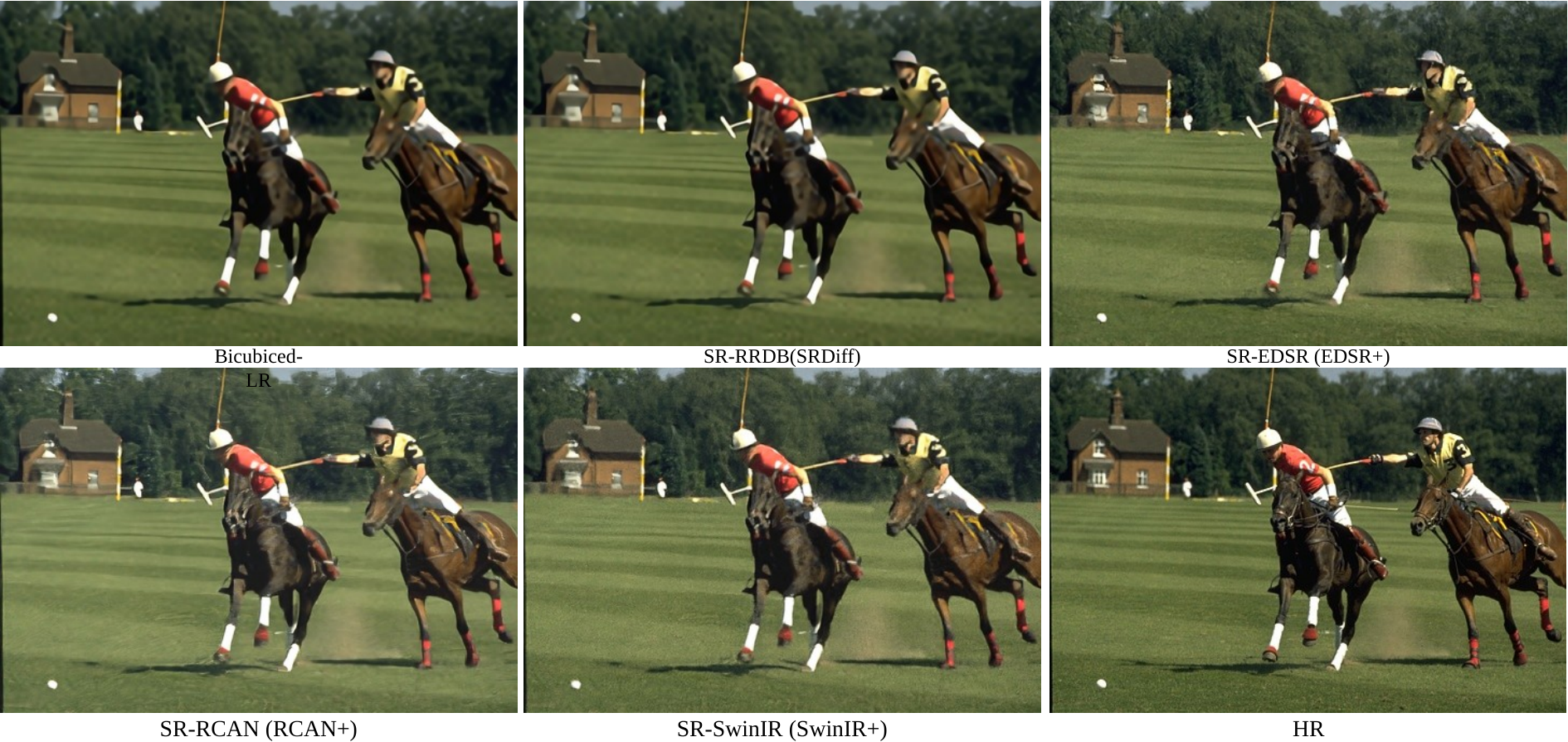}
    \caption{Visual results on '361010' from BSD100 under different conditional images ($\times$4 scale. Best view with zoomed-in.)
    }
    \label{fig:different_conditional_inputs}
\end{figure*}

\textbf{Different conditional images.}
Here, we conduct experiments to verify how different conditional images influence performance. We adopt LR, SRs generated by EDSR, RCAN, SwinIR, and the RRDB trained in SRDiff~\cite{li2022srdiff} as the conditional images to perform experiments, respectively. As shown in Tab.\ref{tab:different_conditional_inputs} and Fig.\ref{fig:different_conditional_inputs}, without any pre-processing, the result under LR conditional performs worst in both quantitative and qualitative. After being pre-trained by RRDB, EDSR, RCAN, and SwinIR, the conditional images can restore more details, pushing our ACDMSR model to perform better.

\begin{table}[!htbp]
  \begin{center}
    \caption{Results of ablation study for different conditional images on BSD100. (4 $\times$ SR)}
  \label{tab:different_conditional_inputs}
   \resizebox{0.9\hsize}{!}{
\begin{tabular}{c|c|c|c|c|c}
\toprule
\hline
% Method & LR     & SR-RRDB & SR-EDSR(EDSR+) & SR-RCAN(RCAN+) & SR-SwinIR(SwinIR+) \\ \midrule
\multirow{2}{*}{Method} & \multirow{2}{*}{LR}   & \multirow{2}{*}{SR-RRDB} &  \multicolumn{3}{c}{\textbf{ours}} \\ \cline{4-6}
&&& \textbf{EDSR+} & \textbf{RCAN+} & \textbf{SwinIR+}\\\midrule
LPIPS$\downarrow$  & 0.1412 & 0.1096  & 0.0935  & 0.0953  &\textbf{ 0.0834}    \\ \hline
PSNR$\uparrow$   & 24.353  & 25.208   & 25.743   & 25.865   &\textbf{ 25.947 }    \\ \hline
SSIM$\uparrow$    & 0.6402 & 0.6589  & 0.6681  & 0.6706  &\textbf{ 0.6743 }   \\ \hline
\bottomrule
\end{tabular}}
  \end{center}
\end{table}

\begin{table}[]
\centering
\caption{Results of ablation study for different loss with 4$\times$SR on Urban100.}
\label{tab:different_loss}
\begin{tabular}{cc|c|c|c}
\toprule
\hline
\multicolumn{2}{c|}{Method}                                     & LPIPS  & PSNR   & SSIM   \\ \hline
\multicolumn{1}{c|}{\multirow{2}{*}{EDSR+}}   & Image-predicted & 0.0997 & 25.452 & 0.7649 \\ \cline{2-5} 
\multicolumn{1}{c|}{}                         & Noise-predicted & 0.1106 & 24.866 & 0.7805 \\ \hline
\multicolumn{1}{c|}{\multirow{2}{*}{RCAN+}}   & Image-predicted & 0.0997  & 25.587 & 0.7681 \\ \cline{2-5} 
\multicolumn{1}{c|}{}                         & Noise-predicted & 0.1051 & 25.035 & 0.7820 \\ \hline
\multicolumn{1}{c|}{\multirow{2}{*}{SwinIR+}} & Image-predicted & 0.0934 & 25.852 & 0.7796 \\ \cline{2-5} 
\multicolumn{1}{c|}{}                         & Noise-predicted & 0.1018 & 25.130 & 0.7915 \\ \hline
\midrule
\end{tabular}
\end{table}

\textbf{ Noise-predicted Loss VS. Image-predicted Loss.}
We conduct an experiment on the Urban100 dataset with scale factor 4 to verify whether training the model to predict noise or images can achieve better performance. As shown in Tab.~\ref{tab:different_loss}, the image prediction model can achieve both better distortion metric (PSNR, SSIM) and perceptual quality (LPIPS), 
compared with  SR3~\cite{saharia2022image} and SRdiff~\cite{li2022srdiff}, whose model predicts the added noise. It is because the image-predict model is more likely to learn the distribution of image information, which helps obtain good results for super-resolution reconstruction.

\section{Conclusion}
Our work revisits diffusion models in super-resolution and reveals that taking a pre-super-resolved version for the given LR image as the conditional image can help to achieve a better high-resolution image. Based on this, we propose a simple but non-trivial DPM-based super-resolution post-process framework,~\ie, ACDMSR. By taking a pre-super-resolved version of the given LR image and adapting the standard diffusion models to perform super-resolution, our ACDMSR improves both qualitative and quantitative results and can generate more photo-realistic counterparts for the low-resolution images on benchmark datasets (Set5, Set14, Urban100, BSD100, Manga109). In the future, we will extend our ACDMSR to images with more complex degradation. 

Although our method achieves impressive results in generating high-quality images in single image super-resolution, however, it inherits the natural issue of the diffusion models that require multiple feedforwards to achieve the final output. The recent progress in the research community attempts to resolve this drawback of the diffusion model to shorten it to a single step with promising results~\cite{salimans2022progressive,zheng2022fast,song2023consistency}, which can be beneficial for the SISR framework proposed by our method. In future work, we will focus on accelerating the inference process of diffusion models for image super-resolution.

\ifCLASSOPTIONcaptionsoff
  \newpage
\fi

\newpage

\bibliographystyle{IEEEtran}
\bibliography{mybibfile.bib}

\begin{thebibliography}{10}
\providecommand{\url}[1]{#1}
\csname url@samestyle\endcsname
\providecommand{\newblock}{\relax}
\providecommand{\bibinfo}[2]{#2}
\providecommand{\BIBentrySTDinterwordspacing}{\spaceskip=0pt\relax}
\providecommand{\BIBentryALTinterwordstretchfactor}{4}
\providecommand{\BIBentryALTinterwordspacing}{\spaceskip=\fontdimen2\font plus
\BIBentryALTinterwordstretchfactor\fontdimen3\font minus
  \fontdimen4\font\relax}
\providecommand{\BIBforeignlanguage}[2]{{%
\expandafter\ifx\csname l@#1\endcsname\relax
\typeout{** WARNING: IEEEtran.bst: No hyphenation pattern has been}%
\typeout{** loaded for the language `#1'. Using the pattern for}%
\typeout{** the default language instead.}%
\else
\language=\csname l@#1\endcsname
\fi
#2}}
\providecommand{\BIBdecl}{\relax}
\BIBdecl

\bibitem{9381876}
C.~Yan, Y.~Hao, L.~Li, J.~Yin, A.~Liu, Z.~Mao, Z.~Chen, and X.~Gao,
  ``Task-adaptive attention for image captioning,'' \emph{IEEE Transactions on
  Circuits and Systems for Video Technology}, 2022.

\bibitem{chen2022consistent}
Z.~Chen, K.-Y. Lin, and W.-S. Zheng, ``Consistent intra-video contrastive
  learning with asynchronous long-term memory bank,'' \emph{IEEE Transactions
  on Circuits and Systems for Video Technology}, 2022.

\bibitem{pham2022self}
T.~X. Pham, A.~Niu, Z.~Kang, S.~R. Madjid, J.~W. Hong, D.~Kim, J.~T.~J. Tee,
  and C.~D. Yoo, ``Self-supervised visual representation learning via residual
  momentum,'' \emph{arXiv preprint arXiv:2211.09861}, 2022.

\bibitem{niu2022fast}
A.~Niu, K.~Zhang, C.~Zhang, C.~Zhang, I.~S. Kweon, C.~D. Yoo, and Y.~Zhang,
  ``Fast adversarial training with noise augmentation: A unified perspective on
  randstart and gradalign,'' \emph{arXiv preprint arXiv:2202.05488}, 2022.

\bibitem{pan2020unsupervised}
F.~Pan, I.~Shin, F.~Rameau, S.~Lee, and I.~S. Kweon, ``Unsupervised
  intra-domain adaptation for semantic segmentation through self-supervision,''
  in \emph{CVPR}, 2020.

\bibitem{pan2022ml}
F.~Pan, S.~Hur, S.~Lee, J.~Kim, and I.~S. Kweon, ``Ml-bpm: Multi-teacher
  learning with bidirectional photometric mixing for open compound domain
  adaptation in semantic segmentation,'' in \emph{ECCV}, 2022.

\bibitem{pan2022labeling}
F.~Pan, F.~Rameau, and I.~S. Kweon, ``Labeling where adapting fails:
  Cross-domain semantic segmentation with point supervision via active
  selection,'' \emph{arXiv preprint arXiv:2206.00181}, 2022.

\bibitem{chang2021two}
K.~Chang, H.~Li, Y.~Tan, P.~L.~K. Ding, and B.~Li, ``A two-stage convolutional
  neural network for joint demosaicking and super-resolution,'' \emph{IEEE
  Transactions on Circuits and Systems for Video Technology}, 2021.

\bibitem{niu2022ms2net}
A.~Niu, Y.~Zhu, C.~Zhang, J.~Sun, P.~Wang, I.~S. Kweon, and Y.~Zhang, ``Ms2net:
  Multi-scale and multi-stage feature fusion for blurred image
  super-resolution,'' \emph{IEEE Transactions on Circuits and Systems for Video
  Technology}, 2022.

\bibitem{niu2023cdpmsr}
A.~Niu, K.~Zhang, T.~X. Pham, J.~Sun, Y.~Zhu, I.~S. Kweon, and Y.~Zhang,
  ``Cdpmsr: Conditional diffusion probabilistic models for single image
  super-resolution,'' \emph{arXiv preprint arXiv:2302.12831}, 2023.

\bibitem{ledig2017photo}
C.~Ledig, L.~Theis, F.~Husz{\'a}r, J.~Caballero, A.~Cunningham, A.~Acosta,
  A.~Aitken, A.~Tejani, J.~Totz, Z.~Wang \emph{et~al.}, ``Photo-realistic
  single image super-resolution using a generative adversarial network,'' in
  \emph{CVPR}, 2017.

\bibitem{zhang2018image}
Y.~Zhang, K.~Li, K.~Li, L.~Wang, B.~Zhong, and Y.~Fu, ``Image super-resolution
  using very deep residual channel attention networks,'' in \emph{ECCV}, 2018.

\bibitem{ma2019matrix}
H.~Ma, X.~Chu, B.~Zhang, and S.~Wan, ``A matrix-in-matrix neural network for
  image super resolution,'' \emph{arXiv preprint arXiv:1903.07949}, 2019.

\bibitem{cai2019toward}
J.~Cai, H.~Zeng, H.~Yong, Z.~Cao, and L.~Zhang, ``Toward real-world single
  image super-resolution: A new benchmark and a new model,'' in \emph{ICCV},
  2019.

\bibitem{zuo2019multi}
Y.~Zuo, Q.~Wu, Y.~Fang, P.~An, L.~Huang, and Z.~Chen, ``Multi-scale frequency
  reconstruction for guided depth map super-resolution via deep residual
  network,'' \emph{IEEE Transactions on Circuits and Systems for Video
  Technology}, 2019.

\bibitem{liu2020photo}
Z.-S. Liu, W.-C. Siu, and Y.-L. Chan, ``Photo-realistic image super-resolution
  via variational autoencoders,'' \emph{IEEE Transactions on Circuits and
  Systems for video Technology}, 2020.

\bibitem{lyn2020multi}
J.~Lyn, ``Multi-level feature fusion mechanism for single image
  super-resolution,'' \emph{arXiv preprint arXiv:2002.05962}, 2020.

\bibitem{li2021single}
X.~Li and Z.~Chen, ``Single image super-resolution reconstruction based on
  fusion of internal and external features,'' \emph{Multimedia Tools and
  Applications}, 2021.

\bibitem{9897540}
H.~Zhang, Y.~Zhu, J.~Sun, and Y.~Zhang, ``Real-world image super-resolution via
  kernel augmentation and stochastic variation,'' in \emph{ICIP}, 2022.

\bibitem{lim2017enhanced}
B.~Lim, S.~Son, H.~Kim, S.~Nah, and K.~Mu~Lee, ``Enhanced deep residual
  networks for single image super-resolution,'' in \emph{CVPR workshops}, 2017.

\bibitem{hu2019channel}
Y.~Hu, J.~Li, Y.~Huang, and X.~Gao, ``Channel-wise and spatial feature
  modulation network for single image super-resolution,'' \emph{IEEE
  Transactions on Circuits and Systems for Video Technology}, 2019.

\bibitem{li2019filternet}
F.~Li, H.~Bai, and Y.~Zhao, ``Filternet: Adaptive information filtering network
  for accurate and fast image super-resolution,'' \emph{IEEE Transactions on
  Circuits and Systems for Video Technology}, 2019.

\bibitem{niu2023gran}
A.~Niu, P.~Wang, Y.~Zhu, J.~Sun, Q.~Yan, and Y.~Zhang, ``Gran: Ghost residual
  attention network for single image super resolution,'' \emph{Multimedia Tools
  and Applications}, 2023.

\bibitem{deng2015single}
L.-J. Deng, W.~Guo, and T.-Z. Huang, ``Single-image super-resolution via an
  iterative reproducing kernel hilbert space method,'' \emph{IEEE Transactions
  on Circuits and Systems for Video Technology}, 2015.

\bibitem{mei2021image}
Y.~Mei, Y.~Fan, and Y.~Zhou, ``Image super-resolution with non-local sparse
  attention,'' in \emph{CVPR}, 2021.

\bibitem{liang2021swinir}
J.~Liang, J.~Cao, G.~Sun, K.~Zhang, L.~Van~Gool, and R.~Timofte, ``Swinir:
  Image restoration using swin transformer,'' in \emph{ICCV}, 2021.

\bibitem{lu2022transformer}
Z.~Lu, J.~Li, H.~Liu, C.~Huang, L.~Zhang, and T.~Zeng, ``Transformer for single
  image super-resolution,'' in \emph{CVPR workshop}, 2022.

\bibitem{wu2021practical}
G.~Wu, J.~Jiang, X.~Liu, and J.~Ma, ``A practical contrastive learning
  framework for single image super-resolution,'' \emph{arXiv preprint
  arXiv:2111.13924}, 2021.

\bibitem{zhu2022self}
Y.~Zhu, H.~Shuai, G.~Liu, and Q.~Liu, ``Self-supervised video representation
  learning using improved instance-wise contrastive learning and deep
  clustering,'' \emph{IEEE Transactions on Circuits and Systems for Video
  Technology}, 2022.

\bibitem{niu2023learning}
A.~Niu, K.~Zhang, T.~X. Pham, P.~Wang, J.~Sun, I.~S. Kweon, and Y.~Zhang,
  ``Learning from multi-perception features for real-word image
  super-resolution,'' \emph{arXiv preprint arXiv:2305.18547}, 2023.

\bibitem{blau2018perception}
Y.~Blau and T.~Michaeli, ``The perception-distortion tradeoff,'' in
  \emph{CVPR}, 2018.

\bibitem{he2018cascaded}
Z.~He, S.~Tang, J.~Yang, Y.~Cao, M.~Y. Yang, and Y.~Cao, ``Cascaded deep
  networks with multiple receptive fields for infrared image
  super-resolution,'' \emph{IEEE transactions on circuits and systems for video
  technology}, 2018.

\bibitem{delbracio2020projected}
M.~Delbracio, H.~Talebi, and P.~Milanfar, ``Projected distribution loss for
  image enhancement,'' \emph{arXiv preprint arXiv:2012.09289}, 2020.

\bibitem{freirich2021theory}
D.~Freirich, T.~Michaeli, and R.~Meir, ``A theory of the distortion-perception
  tradeoff in wasserstein space,'' \emph{NeurIPS}, 2021.

\bibitem{whang2022deblurring}
J.~Whang, M.~Delbracio, H.~Talebi, C.~Saharia, A.~G. Dimakis, and P.~Milanfar,
  ``Deblurring via stochastic refinement,'' in \emph{CVPR}, 2022.

\bibitem{wang2018esrgan}
X.~Wang, K.~Yu, S.~Wu, J.~Gu, Y.~Liu, C.~Dong, Y.~Qiao, and C.~Change~Loy,
  ``Esrgan: Enhanced super-resolution generative adversarial networks,'' in
  \emph{ECCV workshops}, 2018.

\bibitem{soh2019natural}
J.~W. Soh, G.~Y. Park, J.~Jo, and N.~I. Cho, ``Natural and realistic single
  image super-resolution with explicit natural manifold discrimination,'' in
  \emph{CVPR}, 2019.

\bibitem{tian2022generative}
C.~Tian, X.~Zhang, J.~C.-W. Lin, W.~Zuo, Y.~Zhang, and C.-W. Lin, ``Generative
  adversarial networks for image super-resolution: A survey,'' \emph{arXiv
  preprint arXiv:2204.13620}, 2022.

\bibitem{lugmayr2020srflow}
A.~Lugmayr, M.~Danelljan, L.~V. Gool, and R.~Timofte, ``Srflow: Learning the
  super-resolution space with normalizing flow,'' in \emph{ECCV}, 2020.

\bibitem{liang2021hierarchical}
J.~Liang, A.~Lugmayr, K.~Zhang, M.~Danelljan, L.~Van~Gool, and R.~Timofte,
  ``Hierarchical conditional flow: A unified framework for image
  super-resolution and image rescaling,'' in \emph{ICCV}, 2021.

\bibitem{wolf2021deflow}
V.~Wolf, A.~Lugmayr, M.~Danelljan, L.~Van~Gool, and R.~Timofte, ``Deflow:
  Learning complex image degradations from unpaired data with conditional
  flows,'' in \emph{CVPR}, 2021.

\bibitem{bell2019blind}
S.~Bell-Kligler, A.~Shocher, and M.~Irani, ``Blind super-resolution kernel
  estimation using an internal-gan,'' \emph{NeurIPS}, 2019.

\bibitem{emad2021dualsr}
M.~Emad, M.~Peemen, and H.~Corporaal, ``Dualsr: Zero-shot dual learning for
  real-world super-resolution,'' in \emph{WACV}, 2021.

\bibitem{metz2016unrolled}
L.~Metz, B.~Poole, D.~Pfau, and J.~Sohl-Dickstein, ``Unrolled generative
  adversarial networks,'' \emph{arXiv preprint arXiv:1611.02163}, 2016.

\bibitem{ravuri2019classification}
S.~Ravuri and O.~Vinyals, ``Classification accuracy score for conditional
  generative models,'' \emph{NeurIPS}, 2019.

\bibitem{huang2020fast}
L.~Huang and Y.~Xia, ``Fast blind image super resolution using matrix-variable
  optimization,'' \emph{IEEE Transactions on Circuits and Systems for Video
  Technology}, 2020.

\bibitem{lugmayr2021ntire}
A.~Lugmayr, M.~Danelljan, and R.~Timofte, ``Ntire 2021 learning the
  super-resolution space challenge,'' in \emph{CVPR}, 2021.

\bibitem{saharia2022image}
C.~Saharia, J.~Ho, W.~Chan, T.~Salimans, D.~J. Fleet, and M.~Norouzi, ``Image
  super-resolution via iterative refinement,'' \emph{TPAMI}, 2022.

\bibitem{laroche2022bridging}
C.~Laroche and M.~Tassano, ``Bridging the domain gap in real world
  super-resolution,'' in \emph{ICIP}, 2022.

\bibitem{ho2020denoising}
J.~Ho, A.~Jain, and P.~Abbeel, ``Denoising diffusion probabilistic models,''
  \emph{NeurIPS}, 2020.

\bibitem{li2022srdiff}
H.~Li, Y.~Yang, M.~Chang, S.~Chen, H.~Feng, Z.~Xu, Q.~Li, and Y.~Chen,
  ``Srdiff: Single image super-resolution with diffusion probabilistic
  models,'' \emph{Neurocomputing}, 2022.

\bibitem{he2016deep}
K.~He, X.~Zhang, S.~Ren, and J.~Sun, ``Deep residual learning for image
  recognition,'' in \emph{CVPR}, 2016.

\bibitem{zhu2021lightweight}
X.~Zhu, K.~Guo, S.~Ren, B.~Hu, M.~Hu, and H.~Fang, ``Lightweight image
  super-resolution with expectation-maximization attention mechanism,''
  \emph{IEEE Transactions on Circuits and Systems for Video Technology}, 2021.

\bibitem{woo2018cbam}
S.~Woo, J.~Park, J.-Y. Lee, and I.~S. Kweon, ``Cbam: Convolutional block
  attention module,'' in \emph{ECCV)}, 2018.

\bibitem{li2022blueprint}
Z.~Li, Y.~Liu, X.~Chen, H.~Cai, J.~Gu, Y.~Qiao, and C.~Dong, ``Blueprint
  separable residual network for efficient image super-resolution,'' in
  \emph{CVPR}, 2022.

\bibitem{song2020denoising}
J.~Song, C.~Meng, and S.~Ermon, ``Denoising diffusion implicit models,''
  \emph{arXiv preprint arXiv:2010.02502}, 2020.

\bibitem{bansal2022cold}
A.~Bansal, E.~Borgnia, H.-M. Chu, J.~S. Li, H.~Kazemi, F.~Huang, M.~Goldblum,
  J.~Geiping, and T.~Goldstein, ``Cold diffusion: Inverting arbitrary image
  transforms without noise,'' \emph{arXiv preprint arXiv:2208.09392}, 2022.

\bibitem{karras2022elucidating}
T.~Karras, M.~Aittala, T.~Aila, and S.~Laine, ``Elucidating the design space of
  diffusion-based generative models,'' \emph{arXiv preprint arXiv:2206.00364},
  2022.

\bibitem{salimans2022progressive}
T.~Salimans and J.~Ho, ``Progressive distillation for fast sampling of
  diffusion models,'' \emph{arXiv preprint arXiv:2202.00512}, 2022.

\bibitem{song2023consistency}
Y.~Song, P.~Dhariwal, M.~Chen, and I.~Sutskever, ``Consistency models,''
  \emph{arXiv preprint arXiv:2303.01469}, 2023.

\bibitem{nikankin2022sinfusion}
Y.~Nikankin, N.~Haim, and M.~Irani, ``Sinfusion: Training diffusion models on a
  single image or video,'' \emph{arXiv preprint arXiv:2211.11743}, 2022.

\bibitem{lu2022dpm}
C.~Lu, Y.~Zhou, F.~Bao, J.~Chen, C.~Li, and J.~Zhu, ``Dpm-solver: A fast ode
  solver for diffusion probabilistic model sampling in around 10 steps,''
  \emph{arXiv preprint arXiv:2206.00927}, 2022.

\bibitem{kingma2021variational}
D.~Kingma, T.~Salimans, B.~Poole, and J.~Ho, ``Variational diffusion models,''
  \emph{NeurIPS}, 2021.

\bibitem{song2020score}
Y.~Song, J.~Sohl-Dickstein, D.~P. Kingma, A.~Kumar, S.~Ermon, and B.~Poole,
  ``Score-based generative modeling through stochastic differential
  equations,'' \emph{arXiv preprint arXiv:2011.13456}, 2020.

\bibitem{zhang2018unreasonable}
R.~Zhang, P.~Isola, A.~A. Efros, E.~Shechtman, and O.~Wang, ``The unreasonable
  effectiveness of deep features as a perceptual metric,'' in \emph{CVPR},
  2018.

\bibitem{mittal2012making}
A.~Mittal, R.~Soundararajan, and A.~C. Bovik, ``Making a “completely blind”
  image quality analyzer,'' \emph{IEEE Signal processing letters}, 2012.

\bibitem{zheng2022fast}
H.~Zheng, W.~Nie, A.~Vahdat, K.~Azizzadenesheli, and A.~Anandkumar, ``Fast
  sampling of diffusion models via operator learning,'' \emph{arXiv preprint
  arXiv:2211.13449}, 2022.

\end{thebibliography}

\vspace{-40pt}
\begin{IEEEbiography}
[{\includegraphics[width=1in,height=1.25in,clip,keepaspectratio]{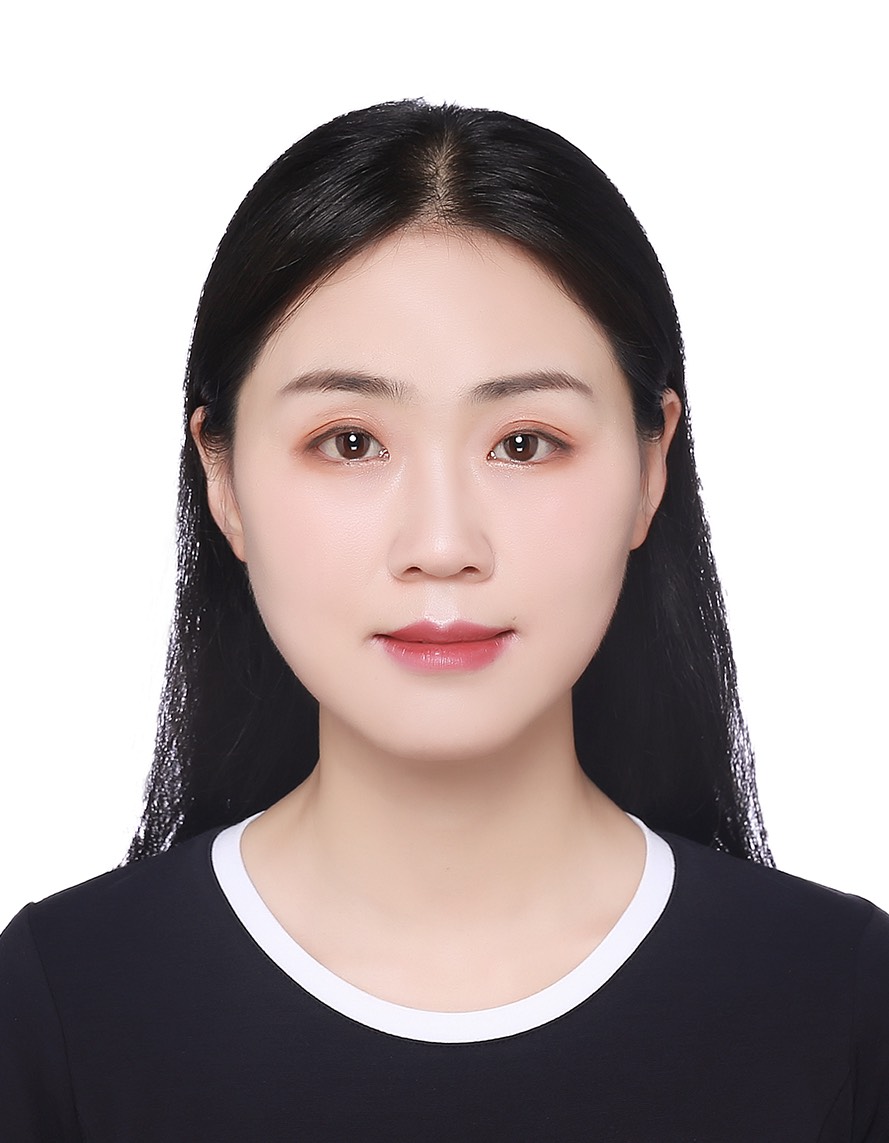}}]
{Axi Niu} received her B.S. and M.S. degrees from Henan  University, Kaifeng, China,
in 2014 and 2017. She is currently pursuing the Ph.D. degree with the School of Computer
Science, Northwestern Polytechnical University, Xi’an, China. Her research interests
include image processing and computer vision.
\end{IEEEbiography}

\vspace{-40pt}
\begin{IEEEbiography}
[{\includegraphics[width=1in,height=1.25in, clip,keepaspectratio]{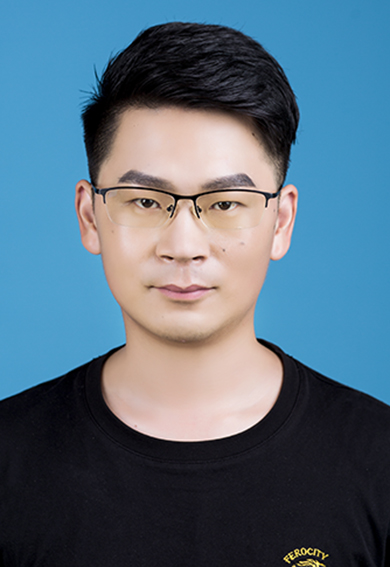}}] {Kang Zhang} received his B.S. degree from Harbin Institute of Technology, 2020. He is currently pursuing the Ph.D. degree at Korea Advanced Institute of Science \& Technology. His research work focuses on Deep Learning, Self-Supervised Learning, and Adversarial Machine Learning. 
\end{IEEEbiography}
\vspace{-30pt}

\begin{IEEEbiography}
[{\includegraphics[width=1in,height=1.25in,clip,keepaspectratio]{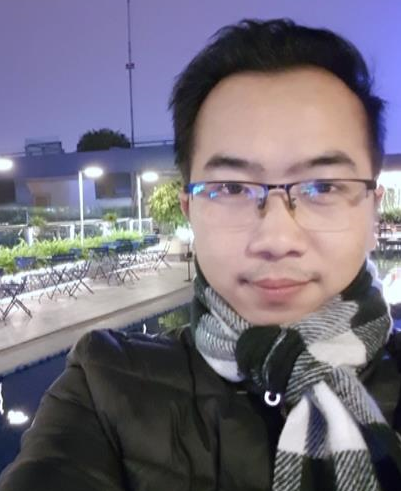}}] { Pham Xuan Trung} received his B.S. degree in the School of Electronics and Telecommunications (SET) at Hanoi University of Science and Technology (HUST) in 2014. He is currently working toward his Ph.D. at KAIST under the supervision of Prof. Chang D. Yoo. His doctoral
research interests include Speech Processing, SelfSupervised Learning, and Computer Vision.
\end{IEEEbiography}
\vspace{-30pt}

\begin{IEEEbiography}
[{\includegraphics[width=1in,height=1.25in,clip,keepaspectratio]{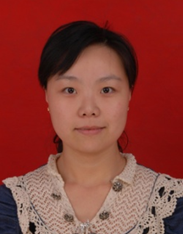}}]
{Jinqiu Sun} received her B.S., M.S., and Ph.D. degrees from Northwestern Polytechnical University in 1999, 2004, and 2005, respectively. She is presently a Professor of the School of Astronomy at Northwestern Polytechnical University. Her research work focuses on signal and image processing, computer vision, and pattern recognition.
\end{IEEEbiography}
\vspace{-30pt}

\begin{IEEEbiography}
[{\includegraphics[width=1in,height=1.25in,clip,keepaspectratio]{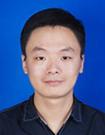}}]
{Yu Zhu} received the B.S., M.S., and M.S. degrees from Northwestern Polytechnical University, Xi’an, China, in 2008, 2011, and 2017, respectively. He is presently an associate researcher at the School of Computer Science, Northwestern Polytechnical University. His current research interests include image processing and image super-resolution.
\end{IEEEbiography}

\begin{IEEEbiography}
[{\includegraphics[width=1in,height=1.25in,clip,keepaspectratio]{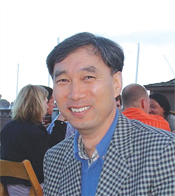}}]
{In So Kweon} received the B.S. and the M.S. degrees in Mechanical Design and Production Engineering from Seoul National University, Korea, in 1981 and 1983, respectively, and the Ph.D. degree in Robotics from the Robotics Institute at Carnegie Mellon University in 1990.  He is currently a Professor of electrical engineering (EE) and the director of the National Core Research Center – P3 DigiCar Center at KAIST. He served as the department head of Automation and Design Engineering (ADE) at KAIST in 1995-1998. His research interests include computer vision and robotics. He has co-authored several books, including "Metric Invariants for Camera Calibration," and more than 300 technical papers. He served as a Founding Associate-Editor-in-Chief for “The International Journal of Computer Vision and Applications”, and has been an Editorial Board Member for “The International Journal of Computer Vision” since 2005. He is  a member of many computer vision and robotics conference program committees and has been a program co-chair for several conferences and workshops. Most recently, he has been a general co-chair of the 2012 Asian Conference on Computer Vision (ACCV) Conference. He received several awards from international conferences, including “The Best Student Paper Runnerup Award in the IEEE-CVPR’2009” and “The Student Paper Award in the ICCAS’2008”. He also earned several honors at KAIST, including the 2002 Best Teaching Award in EE. He is a member of KROS, ICROS, and IEEE.
\end{IEEEbiography}

\begin{IEEEbiography}
[{\includegraphics[width=1in,height=1.25in,clip,keepaspectratio]{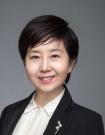}}]
{Yanning Zhang} received her B.S. degree from the Dalian University of Science and Engineering in 1988, M.S. and Ph.D. Degree from Northwestern Polytechnical University in 1993 and 1996, respectively. She is presently a Professor of School of Computer Science and Technology, Northwestern Polytechnical University. She is
also the organization chair of ACCV2009 and the publicity chair of ICME2012. Her research focuses on signal and image processing, computer vision, and pattern recognition. She has published over 200 papers in these fields, including the ICCV2011 best student paper. She is a member of IEEE.
\end{IEEEbiography}

\end{document}